\documentclass[10pt]{article}
\usepackage[margin=.8in]{geometry}

\usepackage{authblk}
\usepackage{graphicx}

\usepackage{amsmath}
\usepackage{hyperref}
\usepackage{xcolor}
\usepackage{amsthm}
\usepackage[normalem]{ulem}
\usepackage{soul}
\usepackage{bbm}
\usepackage{tikz}
\usetikzlibrary{patterns}
\usepackage{caption}
\usepackage{mathrsfs}
\usepackage{xfrac}
\usepackage{wrapfig}
\usepackage[font=small]{caption}
\usepackage{multirow}
\usepackage{algorithm}
\usepackage{algpseudocode}
\usepackage{float}

\usepackage{listings}
\definecolor{commentgreen}{rgb}{0.25,0.5,0.35}
\definecolor{keywordblue}{rgb}{0,0,0.75}
\definecolor{stringred}{rgb}{0.65,0.15,0.15}
\definecolor{identifierblack}{rgb}{0,0,0}
\definecolor{numbergray}{rgb}{0.4,0.4,0.4}
\definecolor{backgroundlightgray}{rgb}{0.97,0.97,0.97}
\usepackage{enumitem}

\usepackage{amsfonts}

\usepackage{cancel}

\usepackage{natbib}
\usepackage[most]{tcolorbox}
\usepackage{xcolor}
\usepackage{setspace}
\tcbuselibrary{listingsutf8}

\newtcolorbox{promptbox}[3][]{
  breakable,
  enhanced,
  listing only,
  listing options={
    basicstyle=\ttfamily\small,
    breaklines=true,
    showstringspaces=false,
    tabsize=4,
    columns=fullflexible,
    keepspaces=true,
    literate={"}{\texttt{"}}{1}
  },
  colback=gray!5!white,
  colframe=black,
  colbacktitle=gray!30!white,
  coltitle=black,
  title=#2,
  fonttitle=\bfseries,
  before upper={%
    \textbf{Description:}~#3\vspace{0.3em}\par
    \rule{\linewidth}{0.4pt}\par
  },
  #1
}

\newtcolorbox{dialoguebox}[3][]{
  breakable,
  enhanced,
  colback=gray!5!white,
  colframe=black,
  colbacktitle=gray!30!white,
  coltitle=black,
  fonttitle=\bfseries,
  title=#2,
  before upper={
    \textbf{Description:}~#3\vspace{0.3em}\par
    \rule{\linewidth}{0.4pt}\vspace{0.5em}\par
  },
  #1
}
\definecolor{patient}{rgb}{0.5,0.1,0.1}
\definecolor{clinician}{rgb}{0.8,0.0,0.3}
\definecolor{prediction}{RGB}{210, 235, 200}

\newcommand{\reasoning}[1]{\colorbox{deliberativeknowledge-light}{\strut\textbf{Clinician Reasons:}} \emph{#1}}
\newcommand{\rag}[1]{\colorbox{documentaryknowledge-light}{\strut\textbf{Clinician Queries Wikipedia:}} \emph{#1}}
\newcommand{\question}[1]{\colorbox{institutionalknowledge-light}{\strut\textbf{Clinician Asks:}} \emph{#1}}
\newcommand{\experiment}[1]{\colorbox{experimentalknowledge-light}{\strut\textbf{Clinician Orders Laboratory Test:}} \emph{#1}}
\newcommand{\environment}[1]{\textbf{Response from $\mathcal{E}$:} \emph{#1}}
\newcommand{\prediction}[1]{\colorbox{prediction}{\strut\textbf{Clinician Predicts:}} \emph{#1}}

\usepackage[toc,page,header]{appendix}
\usepackage{minitoc}

\usepackage{amsmath}
\DeclareMathOperator*{\argmax}{arg\,max}

\newcommand\eg{\textit{e.g.}}

\definecolor{linkblue}{rgb}{0.0, 0.3, 0.6}

\definecolor{directknowledge}{RGB}{242, 140, 140}
\definecolor{deliberativeknowledge}{RGB}{242, 194, 153}
\definecolor{documentaryknowledge}{RGB}{157, 184, 159}
\definecolor{institutionalknowledge}{RGB}{142, 198, 230}
\definecolor{experimentalknowledge}{RGB}{183, 174, 221}
\definecolor{CuriosiTree}{RGB}{211, 163, 163}

\definecolor{directknowledge-light}{RGB}{247, 180, 180}
\definecolor{deliberativeknowledge-light}{RGB}{247, 215, 189}
\definecolor{documentaryknowledge-light}{RGB}{191, 209, 193}
\definecolor{institutionalknowledge-light}{RGB}{182, 218, 239}
\definecolor{experimentalknowledge-light}{RGB}{208, 202, 233}
\definecolor{CuriosiTree-light}{RGB}{226, 195, 195}

\hypersetup{
  colorlinks = true,
  linkcolor=linkblue,
  citecolor=linkblue,
  urlcolor=linkblue,
  bookmarksopen=true,
  pdfdisplaydoctitle=true
}

\title{The Curious Language Model: Strategic Test-Time Information Acquisition}

\author[, 1, 2, 3]{Michael Cooper\thanks{Correspondence to \href{mailto:coopermj@cs.toronto.edu}{coopermj@cs.toronto.edu}. Work completed while working on internship at Abridge AI, Inc.}}
\author[1]{Rohan Wadhawan}
\author[1]{John Michael Giorgi}
\author[1,4]{Chenhao Tan}
\author[1]{Davis Liang}

\affil[1]{Abridge AI, Inc.}
\affil[2]{University of Toronto, Toronto, ON}
\affil[3]{Vector Institute for Artificial Intelligence, Toronto, ON}
\affil[4]{University of Chicago, Chicago, IL}

\begin{document}
\doparttoc
\setcounter{parttocdepth}{3}
\faketableofcontents

\maketitle

\begin{abstract}

Decision-makers often possess insufficient information to render a confident decision. In these cases, the decision-maker can often undertake actions to acquire the necessary information about the problem at hand, \eg, by consulting knowledgeable authorities or by conducting experiments. Importantly, different levers of information acquisition come with different costs, posing the challenge of selecting the actions that are both informative and cost-effective. In this work, we propose \textsc{CuriosiTree}, a heuristic-based, test-time policy for zero-shot information acquisition in large language models (LLMs). \textsc{CuriosiTree} employs a greedy tree search to estimate the expected information gain of each action and strategically chooses actions based on a balance of anticipated information gain and associated cost. Empirical validation in a clinical diagnosis simulation shows that \textsc{CuriosiTree} enables cost-effective integration of heterogenous sources of information, and outperforms baseline action selection strategies in selecting action sequences that enable accurate diagnosis.\footnote{A software implementation of our method can be found at \href{https://github.com/cooper-mj/CuriosiTree}{this code repository}.}

\end{abstract}

\section{Introduction}

As a motivating example for our work, consider the process by which a clinician arrives at a diagnosis for a patient \citep{ball2015improving}. Based on prior information---such as the patient’s chart or history of previous encounters---the clinician forms an initial hypothesis that takes the form of an implicit probability distribution over a set of plausible diagnoses. If the clinician is sufficiently confident in the most likely diagnosis under this distribution, they proceed to diagnose the patient. If not, the clinician may take information-gathering steps to reduce his or her uncertainty. These might include performing a differential diagnosis \citep{harvey1972differential}: proposing a plausible set of diagnoses, then asking the patient further questions, ordering laboratory tests, consulting relevant case studies, or deductively reasoning about the information already available to distinguish between hypotheses. Each action yields a subsequent hypothesis: a posterior distribution that is the result of applying the accumulated evidence to the previous hypothesis \citep{el1995people}. Once the clinician has obtained a posterior that gives them sufficient confidence in the most likely diagnosis, they proceed to render the diagnosis.

Choosing the most informative information-gathering to render diagnosis for a given patient presents a challenging inference problem. Complicating matters further, however, is that these information-gathering actions are not without cost: laboratory tests may carry financial burdens \citep{ji2024measurement}, and reviewing the literature or reflecting on complex cases consumes time. And, because these costs are often \textit{opportunity costs}---meaning they are indirectly borne by the entire patient population \citep{wieser1924theorie, fuertes2017opportunity}---it is essential to choose actions that enable correct diagnosis while incurring minimal cost.

This work proposes a principled, decision-theoretic framework that enables modern large language models (LLMs) to autonomously select information-gathering actions under uncertainty. At each time step, the framework either issues a prediction---if the model estimates high confidence in its correctness based on currently available information---or chooses an information-gathering action aimed at maximally reducing uncertainty.

\begin{wrapfigure}{r}{0.55\textwidth}
    \includegraphics[width=0.55\textwidth]{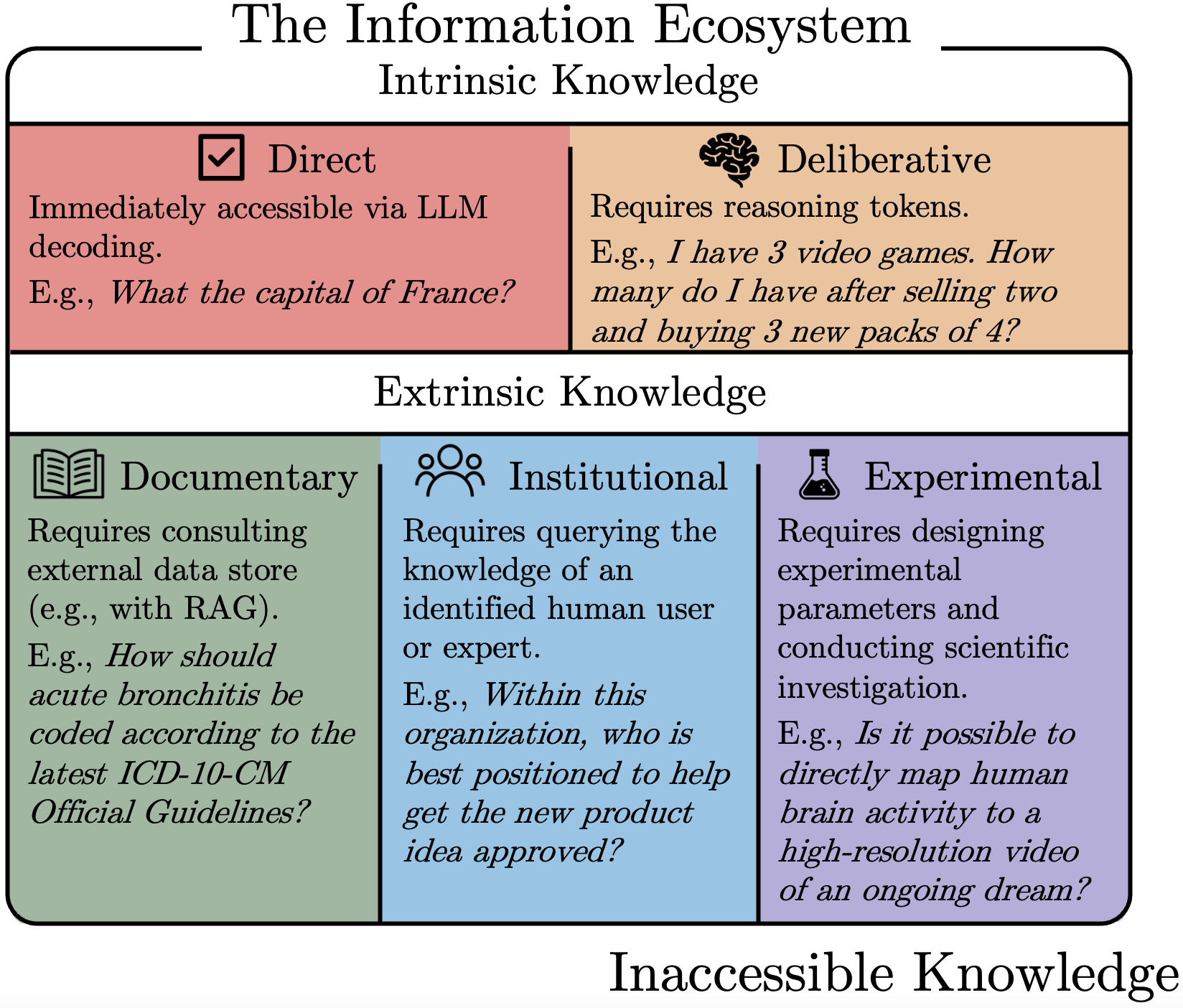}
    \caption{Summary of the different partitions of the Information Ecosystem, and example queries of knowledge that may fall into each partition for some $\mathcal{M}$. We argue that any knowledge not falling within the above five partitions is inaccessible to modern language models, even if augmented with an embodied agent capable of running empirical experiments.}
    \label{fig:info-ecosystem}
\end{wrapfigure}

\subsection{The Information Ecosystem}

We begin by introducing the Information Ecosystem as comprising five distinct partitions, and propose that different kinds of information-gathering actions are required to access knowledge within each partition. We first distinguish between \textit{intrinsic} and \textit{extrinsic knowledge} with respect to a model $\mathcal{M}$, where intrinsic knowledge is the set of queries that $\mathcal{M}$ can resolve using only its own weights, while extrinsic knowledge is the set of queries for which $\mathcal{M}$ requires access to external resources such as a data store, human expert, or embodied agent within the physical world.

Intrinsic knowledge can be either \colorbox{directknowledge-light}{\strut\textit{direct}} or \colorbox{deliberativeknowledge-light}{\strut\textit{deliberative}}: direct knowledge refers to accessible knowledge through standard zero-shot prompting, while deliberative knowledge refers to that which requires generation of intermediary reasoning tokens in order to access (\eg $\,$ \citep{wei2022chain, yao2023tree}). Similarly, extrinsic knowledge can be either \colorbox{documentaryknowledge-light}{\strut\textit{documentary}}, \colorbox{institutionalknowledge-light}{\strut\textit{institutional}}, or  \colorbox{experimentalknowledge-light}{\strut\textit{experimental}}. Documentary knowledge lies in existing textual documents that can be readily indexed, so can be accessed by equipping $\mathcal{M}$ with a retrieval algorithm like retrieval-augmented generation (RAG) \citep{lewis2020retrieval}. Institutional knowledge, in contrast, refers to contextual knowledge contained within the minds of human experts that has not been converted into written form \citep{coffey2003knowledge}. That this knowledge is absent from the textual corpora leveraged by language models means that institutional knowledge can only be accessed by interactively querying the relevant human expert(s). Finally, experimental knowledge refers to that which neither resides in accessible written documents, nor is known to human experts. This represents the frontier of testable knowledge, only accessible via the design and execution of novel empirical investigation (\eg, \citep{lu2024ai}). Figure \ref{fig:info-ecosystem} summarizes the different partitions of the Information Ecosystem and provides example queries of knowledge in each partition.

In this framework, the composition of each partition is a function of $\mathcal{M}$: for example, information about the symptoms of different kinds of heart disease may be considered extrinsic (documentary) knowledge under a model $\mathcal{M}$ that has not been trained on much medical data, but may be considered intrinsic (direct or deliberative) knowledge under a model $\mathcal{M}$ that has been trained on such data.

\subsection{Our Contributions}

The central observation underlying our study is that querying each partition of the Information Ecosystem requires a unique kind of query, and that different kinds of queries typically come with different costs. For example, generating reasoning tokens may be inexpensive relative to conducting a novel empirical investigation or identifying and querying a human expert. Though a substantial body of literature exists on retrieving knowledge from individual partitions, we propose a unified, information-theoretic framework for LLMs to navigate myriad forms of knowledge acquisition in complex environments. Specifically, our work contributes the following:

\begin{enumerate}[leftmargin=*]
    \item We present \textsc{CuriosiTree}, a principled framework for language models to autonomously and strategically query diverse information sources at test time. By optimizing the trade-off between the expected information gain of each action with its associated cost, \textsc{CuriosiTree} efficiently acquires information about a target estimand under a constrained acquisition budget.
    \item Unlike approaches that rely on computationally costly fine-tuning or self-play, \textsc{CuriosiTree} performs \textit{zero-shot, test-time} information acquisition. It can therefore be readily integrated into closed-source, API-served language models without requiring access to model internals.
    \item In simulation, \textsc{CuriosiTree} acquires a greater quantity of relevant information at lower cumulative cost than other methods, enabling more effective and cost-efficient information seeking across heterogenous modalities.
\end{enumerate}

\section{Related Work}

\noindent{\bf Active Learning and Optimal Experiment Design.} Like active learning \citep{cohn1996active, gal2017deep, xia2025selection}, we study the problem of optimally acquiring information to improve predictive accuracy; however, unlike the classical setting, which permits downstream model training and assumes a moderate-to-large dataset, we operate at test-time, and study \textit{instance-wise acquisition}, choosing to gather information about each datum rather than learning patterns between data points. In this way, elements of our approach can be viewed through the lens of optimal experimental design \citep{chaloner1995bayesian, rainforth2024modern, piriyakulkij2024doing}, simulating experiments and outcomes and choosing to perform those with the greatest informative value.\\

\noindent{\bf Tool Use in LLMs.} A substantial body of work implements mechanisms of querying individual partitions of the Information Ecosystem to augment the information available to large language models at test time. Methods for generating deliberative reasoning tokens \citep{wei2022chain, yao2023tree}, querying documentary knowledge \citep{lewis2020retrieval, asai2023self, thulke2024climategpt}, asking informative questions (below), and designing experimental parameters \citep{ghafarollahi2024protagents, li2024mlr, roohani2024biodiscoveryagent}. Despite this, there has been comparatively little study of how to optimally balance between query mechanisms---especially those of varying cost. Although much of the foundational work in LLM tool use is built on in-context learning \citep{schick2023toolformer, hu2023avis, gou2023tora}, in-context learning is often insufficient \citep{huang2023metatool}, and that incorporating simulation- and heuristic-based approaches may offer a solution \citep{wang2024llms}.\\

\noindent{\bf Question-Asking in LLMs.} Several recent works have studied the task of asking informative questions using language models. Standard training-time methods have included fine-tuning with preference optimization \citep{mazzaccara2024learning, zhang2024modeling, lee2025good} or self-play \citep{andukuri2024star}; some recent work has sought to benchmark the performance of question-asking LLMs in settings of insufficient information \citep{wang2024learning, li2025questbench, li2025aligning}. One prominent test-time approach is Uncertainty-of-Thoughts, which attempts to ask the most informative binary question by computing a closed-form expression for the expected information gain over a number of candidate options. Although elements of our work conceptually draw inspiration from Uncertainty-of-Thoughts, our contribution substantially extends their framework by providing a means of accommodating a diverse set of possible actions with open-ended consequences (beyond the ``yes/no'' questions that are the primary focus of their study), performing cost-sensitive action selection, and by leveraging self-assessed confidence scores to relax their assumption of a uniform distribution over predictions at each point in time.

\section{Strategic Test-Time Information Acquisition}

\subsection{Problem Setting and Notation}
Consider one datum $(X^{(i)}_D, Y^{(i)}) \in \mathcal{X}_D \times \mathcal{Y}$, where $\mathcal{X}_{D}$ denotes a covariate space of dimension $D$ (\eg, a patient's complete clinical state) and $\mathcal{Y}$ denotes a label space (\eg, the patient's diagnosis). Because we are dealing with $k$-class classification, $\mathcal{Y} = \{1, ..., k\}$\footnote{We do not require that all $k$ classes be enumerated at initialization, which is useful if the set of possible classes is very large or unknown at the outset. This is discussed further in Section \ref{sec:implementation-via-llms}.}. The label $Y^{(i)}$ is determined by an unknown ground-truth mapping $f^*:\mathcal{X}_D\rightarrow\mathcal{Y}$, such that $Y^{(i)}=f^*(X^{(i)}_D)$. In practice, however, we do not have access to $X^{(i)}_D$. Instead, we initially observe only a projection $X^{(i)}_{d_0}\in\mathcal{X}_{d_0}$ onto a potentially lower-dimensional subspace with dimension $d_0\leq D$. In our motivating example, $X^{(i)}_{d_0}$ may represent the patient's chart or a set of lab results: these provide a partial, but incomplete, representation of the patient's complete clinical state $X^{(i)}_D$.

At each discrete time step $t = 0, 1, ...,$ we consider an agent that can undertake an action $a_t \in \mathcal{A}_t$ to potentially reveal additional information about $X^{(i)}_{D}$. The set of valid actions $\mathcal{A}_t$ may vary with time, \eg, a patient may only be eligible for an MRI scan after having undergone preliminary assessment. Actions are chosen according to a policy $\pi : \bigcup_{t=0}^{\infty} \left\{ (\mathcal{X}_{d_{t'}}, \mathcal{A}_{t'}) \right\}_{t'=0, ..., t-1} \cup \{\mathcal{X}_{d_t}\} \rightarrow \bigcup_{t=1}^{\infty}\mathcal{A}_t$ that produces each action $a_t \in \mathcal{A}_t$ conditional on $\mathcal{H}_{t-1} = \{(X_{d_{t'}}, a_{t'})\}_{t'=0, ..., t-1} \cup \{X_{d_{t}}\}$, the history of covariates observed and actions taken prior to $a_t$.\footnote{Our method does not necessarily assume that the information obtained from the environment is completely accurate. In cases of inaccurate or contradictory retrievals, our framework allows the implementing LLM to infer the reliability of the retrieved information and factor that into the confidence scores assigned to its predictions.}

Consequently, under some environment $\mathcal{E} : \bigcup_{t=0}^{\infty} \left\{ (X_{d_{t}}, a_{t}) \right\}_{t=0, ..., \infty} \rightarrow \bigcup_{t=1}^{\infty}\mathcal{X}_{d_t}$, the observed covariates evolve as,
\begin{equation}
    \underbrace{X^{(i)}_{d_0} \xrightarrow{\mathcal{E}(X^{(i)}_{d_0}, \,a_0), \,\, a_0 \sim \pi(\mathcal{H}_0)}}_{t=0} \,\,\underbrace{X^{(i)}_{d_1} \xrightarrow{\mathcal{E}(X^{(i)}_{d_1}, \,a_1), \,\, a_1 \sim \pi(\mathcal{H}_1)}}_{t=1}\,\, \cdots, \quad \text{where} \quad d_0 \leq d_1, ... \leq D. 
    \label{eq:covariate-evolution}
\end{equation}

Each action incurs a known, nonnegative cost given by $c:\bigcup_{t=1}^{\infty}\mathcal{A}_t\rightarrow\mathbb{R}_+$, and the cumulative cost of acquiring features must respect a predefined budget $B>0$, in that
$\sum_{t'=0}^{t-1} c(a_t)\leq B$ for all $t$.

\subsection{Inference Goal: Selective Zero-Shot Prediction}

Our goal is to accurately gather information to accurately predict the label $Y^{(i)}$ in the \textit{zero-shot setting}, utilizing a family of \textit{pre-specified} predictive functions $\mathcal{G} = \{g_{d_t}\}_{t=0, ..., \infty}$. Each function $g_{d_t} : \mathcal{X}_{d_t} \rightarrow \Delta(\mathcal{Y})$ maps a partially observed covariate vector to a point on the probability simplex over $\mathcal{Y}$. That is, for any $X_{d_t}^{(i)} \in \mathcal{X}_{d_t}$, $g_{d_t}(X_{d_t}^{(i)})$ is the predictive distribution given by
\begin{equation}
g_{d_t}(X_{d_t}^{(i)}) = \widehat{\Pr}(Y^{(i)} \mid X_{d_t}^{(i)}).
\label{eq:predictive-distribution}
\end{equation}

This zero-shot setting implicitly makes a regularity assumption over the ground-truth mapping $f^*$, specifically, that for some intermediary dimension $d_t \leq D$, 
the partial observation $X_{d_t}^{(i)}$ is sufficiently informative to support a meaningful approximation of $\widehat{Y}^{(i)}$ via $g_{d_t}$. Put differently, we assume $f^*$ is sufficiently ``simple" that the bottleneck to its approximation under $\mathcal{G}$ lies in the information content of $X_{d_t}^{(i)}$ and not the complexity of the function family to which $f^*$ belongs.

However, because this regularity is not guaranteed---nor is it guaranteed that a sufficiently informative $X^{(i)}_{d_t}$ can be obtained within the budget $B$---we employ \textit{selective prediction}, rendering a prediction for $\widehat{Y}^{(i)}$ if the output of $g_{d_t}$ supports a sufficiently high confidence in its most likely label. Our selective prediction rule renders a prediction when the maximum score assigned to a given class by $g_{d_t}$ exceeds a target threshold for the minimum acceptable predicted probability, $\tau \in [0, 1]$,

\begin{equation}
\widehat{Y}^{(i)} =
\begin{cases}
\argmax\limits_{y \in \mathcal{Y}} g_{d_t}(X_{d_t}^{(i)})_y & \text{if } \max\limits_{y \in \mathcal{Y}} g_{d_t}(X_{d_t}^{(i)})_y \geq \tau, \\
\emptyset & \text{otherwise.}
\end{cases}
\label{eq:selective-classification}
\end{equation}

\begin{figure}
    \centering
    \includegraphics[width=\linewidth]{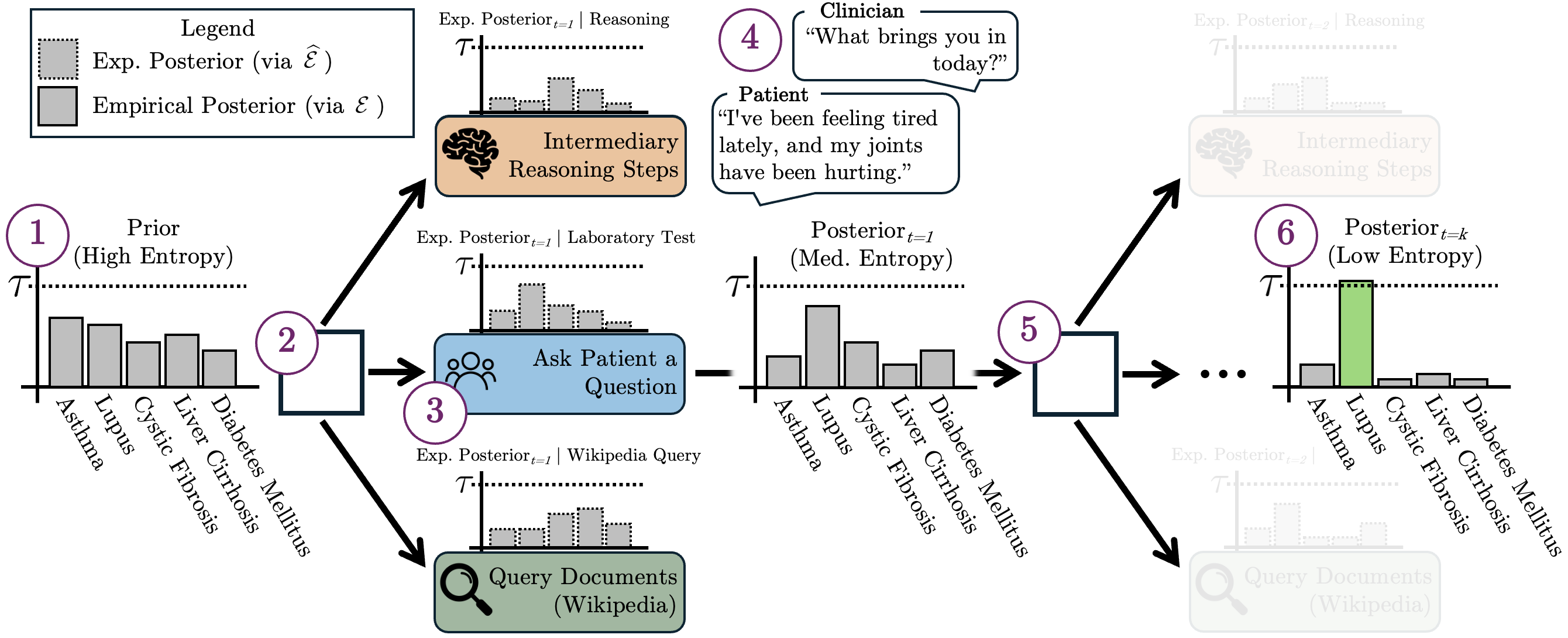}
    \caption{\textsc{CuriosiTree} strategically acquires information at test-time to predict $\widehat{Y}^{(i)}$ within an information acquisition budget. \textcolor{violet}{\bf (1)} $g_{d_0}$ produces a prior distribution (Equation \ref{eq:predictive-distribution}); this distribution has some degree of entropy. \textcolor{violet}{\bf (2)}  $\widehat{\mathcal{E}}$ simulates the effect of taking various different information-gathering actions on our predictive distribution (Equation \ref{eq:infogain}), \textcolor{violet}{\bf (3)} \textsc{CuriosiTree} chooses  to undertake the action with the greatest expected information gain relative to its cost (Equation \ref{eq:action-selection}); in this case, this action entails \colorbox{institutionalknowledge-light}{asking the patient}, ``What brings you in today?'' \textcolor{violet}{\bf (4)} Taking this action in $\mathcal{E}$ yields a posterior distribution via $g_{d_1}$: this distribution concentrates probability mass more tightly around ``lupus,'' based on the new information received. \textcolor{violet}{\bf (5)} Actions with cumulative cost exceeding $B$ are pruned, and the process continues until \textcolor{violet}{\bf (6)} the most probable class under the posterior distribution has a probability score exceeding $\tau$; at this point, \textsc{CuriosiTree} renders a prediction for $\widehat{Y}^{(i)}$ (Equation \ref{eq:selective-classification}). If the most probable class under the predictive posterior distribution does not obtain a score exceeding $\tau$ within the budget $B$, the method abstains. In this way, \textsc{CuriosiTree} implements a test-time procedure that maximally gathers information relevant to predicting $Y^{(i)}$ while minimizing the cumulative cost of the undertaken actions.}
    \label{fig:overview}
\end{figure}

\subsection{\textsc{CuriosiTree}: Strategic Cost-Effective Information Acquisition via Tree Search}

To achieve our inference goal, we seek a policy $\pi$ that balances the information gained under each action with the cost incurred by the action. However, in the zero-shot setting, $\pi$ cannot be learned from data as it is assumed that there are no prior trajectories to draw upon. As such, we implement $\pi$ as a heuristic policy combining principles from decision theory and information theory.

At time $t$, consider the distribution induced over $\mathcal{Y}$ by $g_{d_t}$. If we have access to an \textit{environment simulator}, $\widehat{\mathcal{E}} : \bigcup_{t=0}^{\infty} \left\{ (X_{d_{t}}, a_{t}) \right\}_{t=0, ..., \infty} \rightarrow \bigcup_{t=1}^{\infty}\mathcal{X}_{d_t}$ that predicts the way in which a given action transforms the observed covariates, we can represent the utility of each action using the \textit{expected information gain} (EIG) of each action under $\widehat{\mathcal{E}}$. We write
\begin{align}
    \text{EIG}(a_t|X_{d_{t}}^{(i)}) &= H\left(\widehat{\Pr}(Y^{(i)} \mid X_{d_t}^{(i)})\right) - \mathbb{E}_{\tilde{X}^{(i)}_{d_{t+1}} \sim \widehat{\mathcal{E}}(X_{d_{t}}^{(i)}, a_t)}\left[H\left(\widehat{\Pr}(Y^{(i)} \mid \tilde{X}_{d_{t+1}}^{(i)})\right)\right],
    \label{eq:infogain}
\end{align}

where $H$ is the Shannon entropy \citep{shannon1948mathematical}. We then implement $\pi$ via the following sampling scheme using Equation \ref{eq:infogain} as a guiding heuristic. At time step $t$, observe $X_{d_t}^{(i)}$. Sample $k'$ action candidates, $\{\tilde{a}_t^{(0)}, ..., \tilde{a}_t^{(k'-1)}\} \subseteq \mathcal{A}_t$, then choose $a_t$ to satisfy
\begin{equation}
    a_t = \argmax_{\tilde{a}_t^{(j)} \in \{\tilde{a}_t^{(0)}, ..., \tilde{a}_t^{(k'-1)}\}}  \left[\text{EIG}(\tilde{a}_t^{(j)} | X_{d_t}^{(i)}) - \lambda c(\tilde{a}_t^{(j)})\right],
    \label{eq:action-selection}
\end{equation}

where $\lambda \in \mathbb{R}_+$ is a scaling hyperparameter controlling the tradeoff between utility and cost. In practice, because the expectation in Equation \ref{eq:infogain} is analytically intractable, we compute it using Monte Carlo. A complete algorithm box is provided in the supplementary materials.

\subsection{Implementation via Large Language Models}
\label{sec:implementation-via-llms}

We operationalize our method using large language models \citep{radford2019language}, which are especially appealing in this setting due to their flexibility. This enables us to implement both the family of prediction functions $\mathcal{G}$ and the environment simulator $\widehat{\mathcal{E}}$ using prompting \citep{kojima2022large, hao2023reasoning}. In this setting, $Y^{(i)}$ is a semantic concept, each $X_{d_t}^{(i)}$ is a textual description containing partial information needed to recover $Y^{(i)}$, and $a_t$ is a textual query that induces the progressive update of the sequence of observed covariates as in Equation \ref{eq:covariate-evolution}. All prompts used in our experiments can be found in the supplement. Notably, in our experiments \textit{we prompt} $\widehat{\mathcal{E}}$ \textit{differently than} $\mathcal{E}$, so as to emulate the dynamics of having an imperfect environment simulator, which is often the case in practice.

{\bf Implementing $\widehat{\mathcal{E}}$ via Prompting.} Recall that $\widehat{\mathcal{E}}$ represents the environment simulator for the purpose of evaluating Equation \ref{eq:action-selection}. In theory, the implementing language model should emit responses roughly proportionally consistent with the present set of hypotheses. In practice, the language model emissions are of considerably lower-entropy, which worsens the quality of the approximation. To we circumvent this challenge with a strategy we call \textit{prior locking}, where we explicitly compute the expectation in Equation \ref{eq:infogain} as
\begin{equation}
    \frac{1}{k} \sum_{j=1}^k \mathbb{E}_{\tilde{X}^{(i)}_{d_{t+1}} \sim \widehat{\mathcal{E}}(X_{d_{t}}^{(i)}, a_t, \widehat{y}_{t,j}^{(i)})}\left[H\left(\widehat{\Pr}(Y^{(i)} \mid \tilde{X}_{d_{t+1}}^{(i)})\right)\right],
    \label{eq:priorlocking}
\end{equation}
and explicitly condition each call to the environment simulator by a candidate label $\widehat{y}_{t,j}^{(i)}$, representing one of the top-$k$ model predictions at time $t$.

{\bf Implementing $\mathcal{G}$ via Prompting.} By implementing each $g_{d_t} \in \mathcal{G}$ as a generative process using top-$k$ sampling and constrained decoding \citep{deutsch2019general, kuchnik2023validating}, our method does not assume access to an exhaustive label set. Our prompt for each $g_{d_t}$ accepts as input the observed history of covariates and actions, $\mathcal{H}_t$, and requests a structured list of (a) the $k$ most probable values of the distribution given the current history, $\widehat{y}_{t,0}^{(i)}, ..., \widehat{y}_{t,k-1}^{(i)}$, and (b) the probability scores corresponding to each value, $\tilde{p}_{t,0}^{(i)}, ..., \tilde{p}_{t,k-1}^{(i)}$. Although constrained decoding guarantees that exactly $k$ items and scores are generated, and that the scores are nonnegative floating point numbers, it does not guarantee that the distribution is normalized. We therefore normalize the distribution by dividing each score by the sum of scores assigned across the top-$k$ classes, $\widehat{p}_{t,j}^{(i)}, = \tilde{p}_{t,j}^{(i)} / \sum_{l=0}^{k-1} \tilde{p}_{t,l}^{(i)}$.

This setup makes it challenging to directly evaluate Equation \ref{eq:infogain}, because (a) these top-$k$ distributions are necessarily sparse and unnormalized, meaning we have no closed-form distribution over which we can compute entropy, and (b) because we want Equation \ref{eq:infogain} to reflect the \textit{semantic distribution of candidate labels} in accommodating cases where probability mass is distributed over distinct-but-semantically-similar concepts. We therefore use a surrogate of Equation \ref{eq:infogain} that rewards actions based on the proportion of present candidates that they eliminate from consideration. Using constrained decoding, our prompt to implement this accepts as input the current top-$k$ list of candidates $Y^{(i)}$, the action $a_t$, and the sampled response $\tilde{X}_{d_t+1}^{(i)}$, and produces a structured boolean array $A^{(i)}_t \in [0,1]^k$ of $k$ booleans indicating whether each element in the list of candidates remains logically consistent given the action and the response (1 if \texttt{True}). Then, we implement Equation \ref{eq:action-selection} using $\widehat{\text{EIG}}(a_t|X_{d_t}^{(i)}) = -\log \left[(1/k) \sum_{l=0}^{k-1} A^{(i)}_{t,l}\right]$, with $\widehat{\text{EIG}}(a_t|X_{d_t}^{(i)})$ set to $0$ if no options are consistent. In a sense, this approach implicitly assumes that the true label is among the top-$k$ candidates and rewards actions that are likely to disambiguate among them. Although this approximation introduces potential bias when the ground truth label is not represented among the current predictions, empirically, we find it yields an effective search heuristic.

\section{Experiments}

\subsection{Clinical Diagnosis Simulator}

Clinical diagnosis represents a classic setting for heterogenous information acquisition, as doctors possess a myriad of information-gathering steps that they can undertake to diagnose a patient, each with associated cost. Our primary experimental setting consists of setting $\mathcal{E}$ to be the following clinical diagnosis simulator: a specially-prompted large language model is instantiated with ground-truth underlying diagnosis $Y^{(i)}$, and is tasked with simulating the evolution of the observed patient information in response to a sequence of actions undertaken by the patient's clinician. In our study, this ``clinician'' role is played by an interacting language model, the actions of which are guided by some policy. Our simulator provides the following four levers of information acquisition at each time step. Prompts for the simulator and the interacting model are provided in the supplementary materials.

For the purposes of this testbed, the cost assigned to each class of action is largely arbitrary, though their relative cost-ordering approximately reflects realistic clinical practice. In reality, the exact cost values would be determined by a combination of financial cost (\eg, token generation may be inexpensive; running a laboratory test may be expensive) and other considerations (\eg, limited availability for a certain form of laboratory test).

\begin{enumerate}[leftmargin=*]
    \item \colorbox{deliberativeknowledge-light}{\strut{\bf Generate Reasoning Tokens}} (Cost: 1){\bf .} The clinician can generate intermediary reasoning tokens regarding the patient's condition. These tokens are added to the history so that future predictions and actions are conditional on these tokens. Because reasoning is an intrinsic process, no corresponding response is provided by the experiment simulator.
    \item \colorbox{documentaryknowledge-light}{\strut{\bf Perform RAG on Wikipedia}} (Cost: 1){\bf .}  The clinician can generate a sequence of tokens corresponding to a query for information. In response to the information query, the experiment simulator matches the query to the $p$ most similar chunks of text from Wikipedia using a search over the \href{https://huggingface.co/NeuML/txtai-wikipedia}{\texttt{NeuML/txtai-wikipedia}} embeddings. The text of the chunks is added to the history.
    \item \colorbox{institutionalknowledge-light}{\strut{\bf Ask the Patient a Question}} (Cost: 2){\bf .} The clinician can generate a sequence of tokens representing a question for the patient. In response, the experiment simulator instantiates a language model that has been prompted to respond as a patient whose ground-truth diagnosis is $Y^{(i)}$. This language model is prompted to respond with ignorance if the question asked by the clinician is too technical or specific. The question and response are then added to the history. This baseline is most conceptually similar to an open-ended version of Uncertainty-of-Thoughts.
    \item \colorbox{experimentalknowledge-light}{\strut{\bf Requisition a Laboratory Test}} (Cost: 3){\bf .} The clinician can generate a sequence of tokens representing a specific laboratory assessment to be performed on the patient. In response, the experiment simulator instantiates a language model that has been prompted to yield the results of a laboratory test corresponding to a patient with a ground-truth diagnosis is $Y^{(i)}$. The request and test result are then added to the history.
\end{enumerate} 
No initial information is provided to the clinician at the start of simulation. For each of the four classes of action, we generate five action candidates for a total of twenty actions evaluated each turn. Each action is scored, either using the \textsc{CuriosiTree} heuristic (Equation \ref{eq:action-selection}) or one of the baseline heuristics described below, and the highest-scored action is passed to the environment for evaluation. The simulation is terminated under only two conditions. First, if the cumulative cost incurred exceeds the budget, $B$: we track this as an abstention ($\widehat{Y}^{(i)}=\emptyset$), because insufficient information was gathered to make a confident prediction. Second, if a prediction for $\widehat{Y}^{(i)}$ is rendered according to Equation \ref{eq:selective-classification}: if the prediction is correct (adjudicated using LLM-as-a-judge \citep{zheng2023judging}), we record this as a successful diagnosis, and if the prediction is incorrect, we record this as an unsuccessful diagnosis.

\subsection{Correspondence with Human Intuition}
\label{sec:human-intuition}

\begin{wrapfigure}{r}{0.35\textwidth}
    \includegraphics[width=0.35\textwidth]{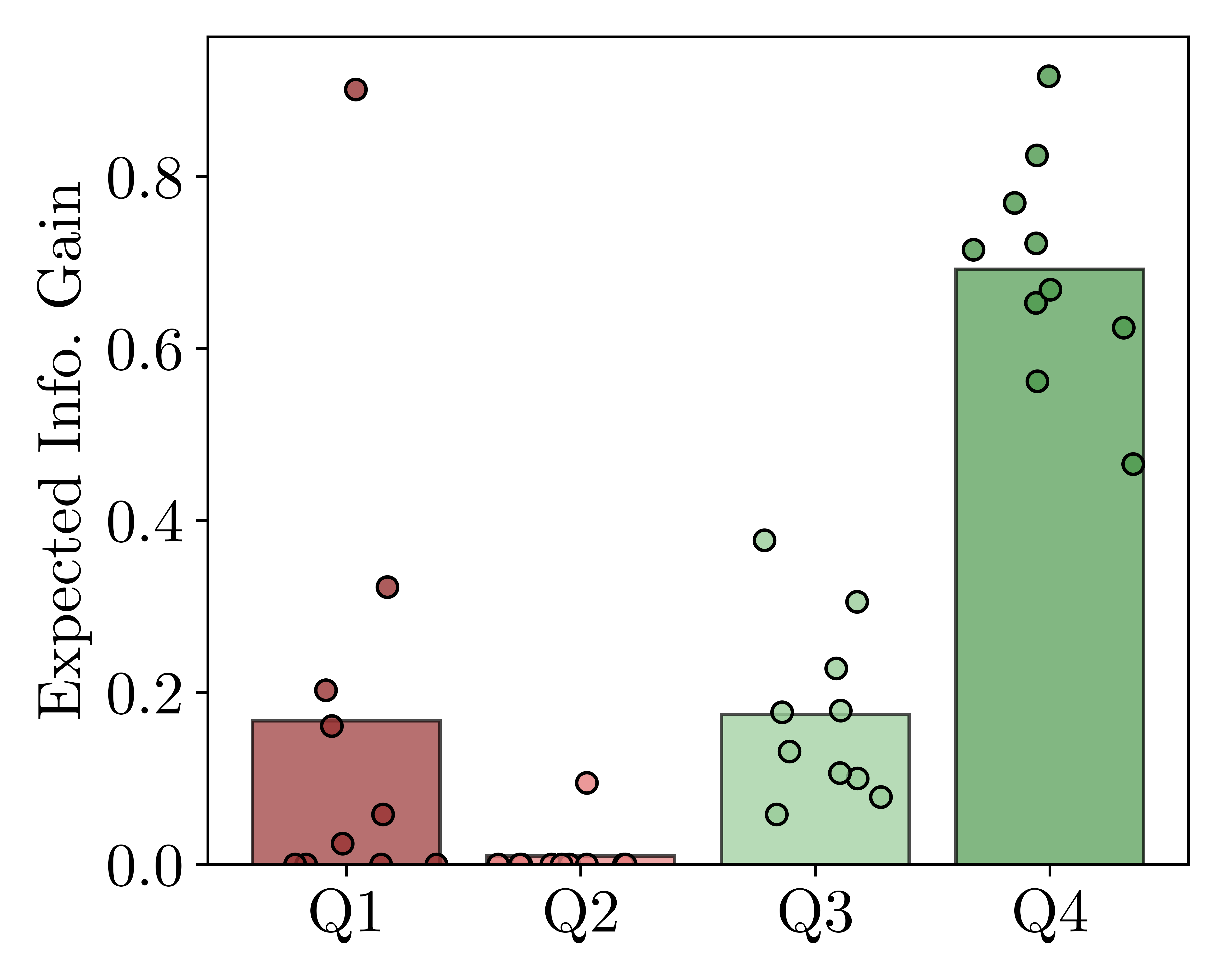}
    \caption{The orderings of Q1 through Q4 (Section \ref{sec:human-intuition}) using the \textsc{CuriosiTree} utility function correspond with human intuition about the value of each question in the clinical diagnostic setting.}
    \label{fig:human-intuition}
\end{wrapfigure}

We present a simple validation of our heuristic by confirming that the utility assigned to four different questions of varying quality correlate with human intuition. We evaluate questions that are irrelevant (\textit{``Q1: What is the best brand of BBQ for an outdoor family event?''}), overly-specific (\textit{``Q2: Is your diagnosis primary biliary cholangitis?''}), weakly-informative (\textit{``Q3: Is the problem associated with your legs?''}), and strongly-informative (\textit{``Q4: What brings you into the clinic today?''}), and we compute EIG as described in Equation \ref{eq:infogain}. We anticipate that the irrelevant and overly-specific questions are assigned low utility scores, while the weakly- and strongly-informative questions enjoy higher utility scores. The result in Figure \ref{fig:human-intuition} confirms that the EIG associated with each question corresponds largely with human intuition.

\begin{figure}[h]
    \centering
    \includegraphics[width=\textwidth]{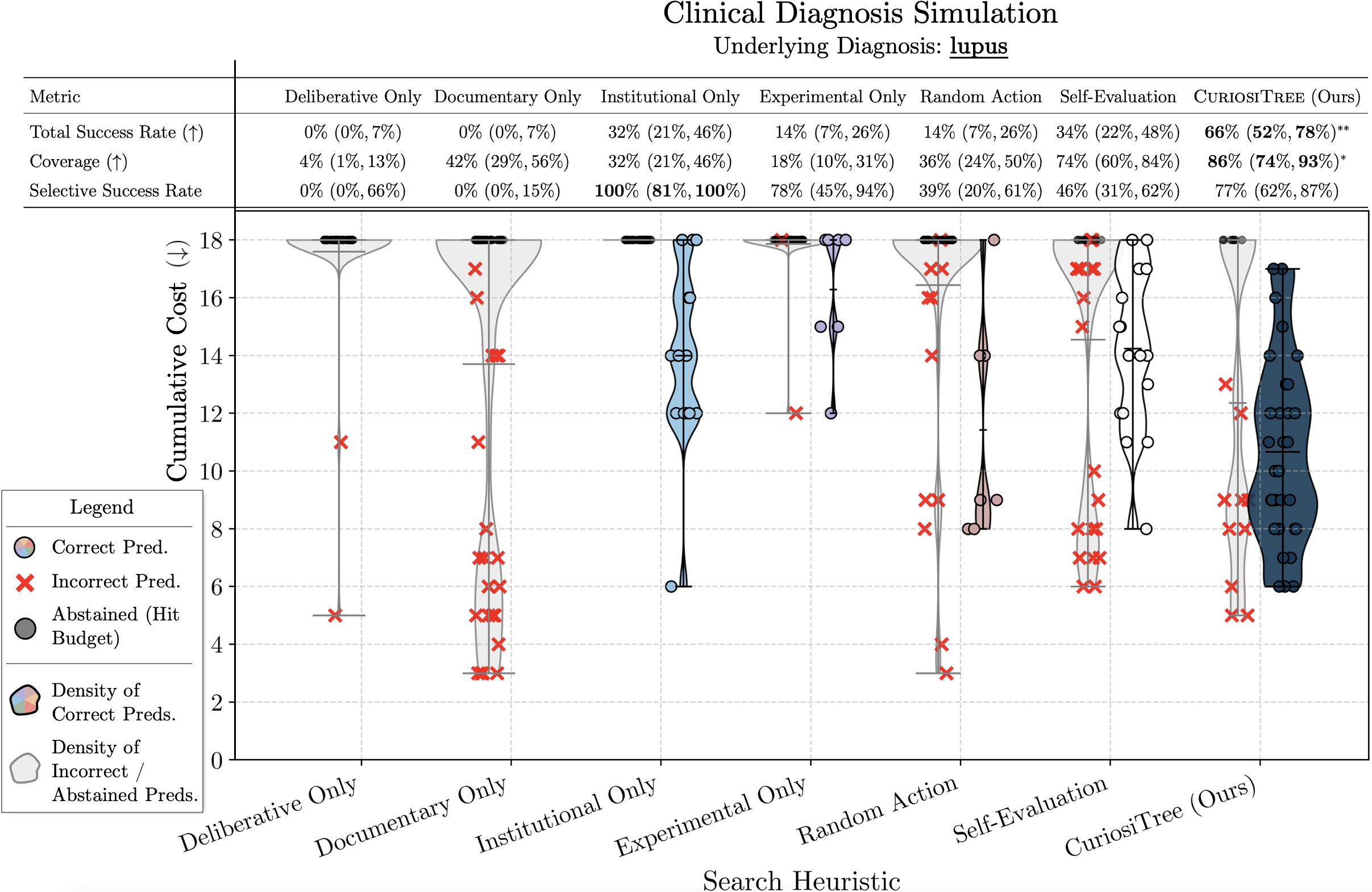}
    \caption{Result of simulation runs with a ground-truth diagnosis ``lupus''. \textit{(Top)} Table highlighting the Total Success Rate, Coverage, and Selective Success Rate of each method. Observe that \textsc{CuriosiTree} enjoys a significantly higher Total Success Rate and Coverage than baseline methods. \textit{(Bottom)} Violin plots representing the cumulative cost incurred by each simulation. For each method, we show two violin plots: the leftmost violin (in gray) represents simulation runs that either rendered an incorrect prediction or encountered the acquisition budget, while the rightmost violin (in colour) represents simulation runs for which the method correctly predicted the patient's diagnosis. A better method is one that corresponds to a lower, fatter, coloured violin, as that represents a method that succeeds more often while incurring less cumulative cost. In this set of fifty simulation trials, \textsc{CuriosiTree} visually achieves higher success at lower cost than the baseline methods.}
    \label{fig:main-result}
\end{figure}

\subsection{Baselines and Experiments}

We implement both \textit{unimodal baselines}---baselines consisting of a single class of action---and \textit{multimodal baselines}---baselines consisting of more than one class of action---to evaluate our method. As one of the key contributions of \textsc{CuriosiTree} is to natively integrate different sources of information into LLM decision-making, the unimodal baselines highlight the importance of consulting heterogenous information sources when making decisions. The multimodal baselines, in constrast, contrast the \textsc{CuriosiTree} heuristic with other plausible heuristics.

Our four unimodal baselines apply the \textsc{CuriosiTree} heuristic to a clinician agent that is capable of selecting actions from a single class: the ``Deliberative Only'' baseline is only capable of generating reasoning tokens; the ``Retrieval Only'' baseline is only capable of performing RAG on Wikipedia; the ``Institutional Only'' baseline is only capable of asking the patient a question; and the ``Experimental Only'' baseline is only capable of requisitioning laboratory tests. Our two multimodal baselines consist of ``Random Action Selection,'' in which an action is selected randomly from the candidate choices, and ``Self-Evaluation,'' in which a prompted language model assigns scores to each action reflecting their informativeness relative to cost. Comparison against the “Random Action Selection” baseline evaluates the benefit of using the \textsc{CuriosiTree} heuristic beyond the language model’s default bias toward contextually plausible actions, while comparison against the “Self-Evaluation” baseline tests whether explicitly optimizing a balance of information gain and cost yields better decision-making than implicit estimates of utility.

We select a series of ten diagnoses spanning varying degrees of commonality and diagnostic difficulty. For each diagnosis, we evaluate each method of information acquisition fifty times. For each set of fifty trials, we track the following summary statistics: Total Success Rate ($\frac{\# \text{ trials where $\widehat{Y}^{(i)} = Y^{(i)}$}}{\# \text{ trials}}$), Coverage ($\frac{\# \text{ trials where $\widehat{Y}^{(i)} \neq \emptyset$}}{\# \text{ trials}}$), and Selective Success Rate ($\frac{\# \text{ trials where $\widehat{Y}^{(i)} = Y^{(i)}$}}{\# \text{ trials where $\widehat{Y}^{(i)} \neq \emptyset$}}$). For brevity, we report only the results for the diagnosis, ``lupus'' in the main manuscript, and report the results on other diagnoses in the supplementary materials. All of our experiments are run using the \texttt{Llama-3.1-70B-Instruct} large language model \citep{grattafiori2024llama}, served using VLLM \citep{kwon2023efficient}. For these experiments, we use a tradeoff parameter of $\lambda = 0.1$.

\section{Results and Discussion}
\label{sec:results-discussion}

\noindent{\bf \textsc{CuriosiTree} Selects More Efficient Information-Gathering Actions.} Figure \ref{fig:main-result} highlights how \textsc{CuriosiTree} enjoys a higher success rate that the baseline methods while incurring lower cost. This suggests that there is value in incorporating formal decision-theoretic heuristics to guide information, rather than prompting the LLM to assess this entirely in-context.\\

\noindent{\bf Integrating Heterogenous Information Sources is Useful and Desirable.} As a simple sanity check, Figure \ref{fig:main-result} demonstrates that intrinsic knowledge alone is insufficient to perform clinical diagnosis: neither the Deliberative Only, nor Documentary Only baselines ever produce the correct diagnosis. In general, incorporating additional data modalities leads to improved overall accuracy, suggesting that our motivating hypothesis---the need to flexibly integrate heterogenous sources of information acquisition---is sound. The one notable exception to this is the Institutional Only baseline, which performs favourably against Random Action selection and Self-Evaluation. Investigating this further, we suggest this is due to the patient response provided by $\mathcal{E}$ being comparatively descriptive, so the institutional knowledge baseline constrains the action space to a uniquely informative subset. We have included sample dialogues in the supplementary materials. In Figure \ref{fig:action-selection}, we see that \textsc{CuriosiTree}---despite its imperfect environment simulator, $\widehat{\mathcal{E}}$---seems to identify the value of Institutional Knowledge in this setting, as queries of Institutional Knowledge represent a plurality of selected actions. Neither of the other multimodal baselines exhibit this behaviour.\\

\noindent{\bf Intrinsic Knowledge May Serve to Sharpen the Predictive Distribution}. We suggest that the primary role of Reasoning steps (intrinsic knowledge) is to sharpen the predictive distribution rather than to alter its most likely prediction. To test this, we identified 95 instances from the experiment above in which \textsc{CuriosiTree} undertook a nonterminal Reasoning step. In 84\% of these cases the most likely diagnosis remained unchanged, indicating that Reasoning steps often preserve the top prediction, and in 62\% of these cases the reasoning step reduced the entropy of the posterior predictive distribution. This suggests that intrinsic Reasoning steps act primarily to sharpen the predictive distribution without altering the its most likely prediction.\\

\noindent{\bf Limitations.} Our study and method are not without limitations. First, assuming a fixed set of $k'$ actions, $k$ elements in the current predictive distribution, and $m$ Monte Carlo steps to compute the expectation in Equation \ref{eq:priorlocking}, \textsc{CuriosiTree} requires model inference be called $O(kk'm)$ times. Despite this, the fact that these calls can be performed in parallel is a redeeming virtue, as its runtime scales favourably with the amount of available compute. Second, despite the strengths of LLM-as-a-judge \citep{zheng2023judging} in emulating human behaviour, our Clinical Diagnosis Simulation may not comprehensively replicate the intricate practicalities of the clinical diagnosis decision-making pipeline.

\begin{figure}[h]
    \centering
    \includegraphics[width=0.95\textwidth]{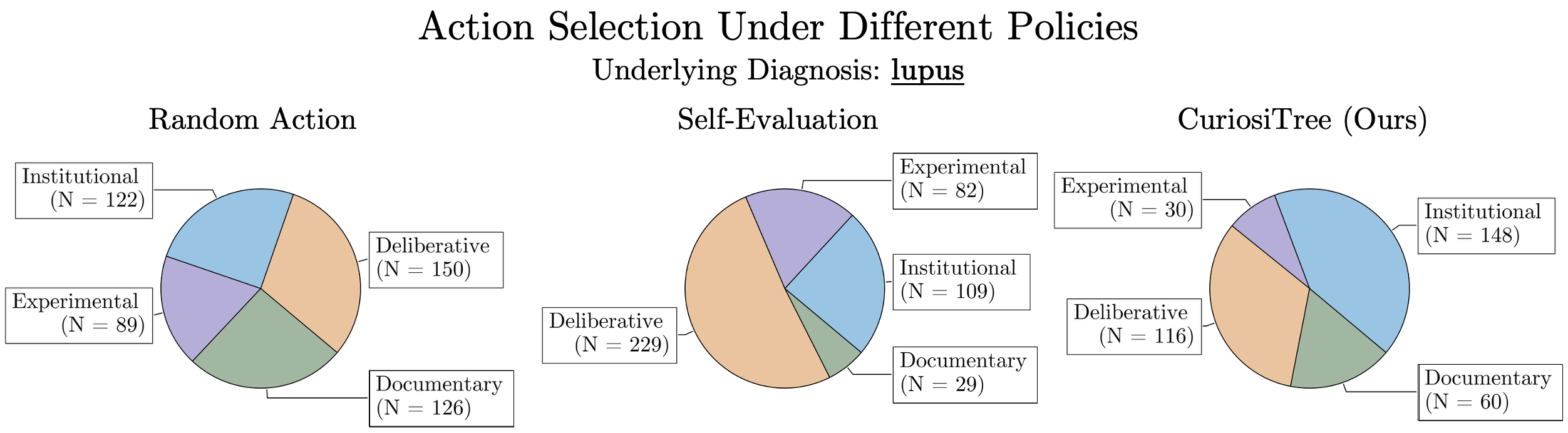}
    \caption{The action classes selected under different policies. Random Action selection induces a more uniform distribution over action classes, while Self-Evaluation is dominated by Deliberative actions. \textsc{CuriosiTree} balances between Institutional and Deliberative actions relatively evenly.}
    \label{fig:action-selection}
\end{figure}

\section{Conclusion}

\textsc{CuriosiTree} imbues language models with the ability to efficiently and autonomously navigate the Information Ecosystem for prediction and decision-making. Our findings highlight the utility of structured, cost-aware exploration in LLM-based agents and open avenues for broader deployment in real-world, resource-constrained settings. We recommend that future work explore the integration of methods like \textsc{CuriosiTree} with more complex data modalities, or in settings with action costs that are unknown or difficult to estimate.

\bibliography{references}

\newpage
\part{Supplementary Material}
\parttoc

\appendix

\section{\textsc{CuriosiTree} Algorithm}

\begin{algorithm}[H]
\caption{\textsc{CuriosiTree}: Strategic, Test-Time Information Acquisition}
\label{alg:curiositree}
\begin{algorithmic}[1]
\Require Initial covariates $X^{(i)}_{d_0}$, predictive functions $\mathcal{G} = \{g_{d_t}\}$, environment simulator $\widehat{\mathcal{E}}$, cost function $c$, prediction threshold $\tau$, budget $B$, utility-cost tradeoff $\lambda$, top-$k$ size $k$, number of action samples $k'$.
\State $C \gets 0$
\State $t \gets 0$
\While{true}
    \State Compute predictive distribution $g_{d_t}(X_{d_t}^{(i)})$ with top-$k$ predictions $\{\widehat{y}_{t,j}^{(i)}, \widehat{p}_{t,j}^{(i)}\}_{j=0}^{k-1}$
    \If{$\max_j \widehat{p}_{t,j}^{(i)} \geq \tau$}
        \State \Return $\widehat{Y}^{(i)} \gets \arg\max_j \widehat{p}_{t,j}^{(i)}$
    \EndIf
    \State Sample $k'$ actions $\{\tilde{a}_t^{(j)}\}_{j=0}^{k'-1} \subseteq \mathcal{A}_t$ that do not exceed the budget: $c(\tilde{a}_t^{(j)}) + C \leq B$
    \For{each candidate action $\tilde{a}_t^{(j)}$}
        \For{each top-$k$ prediction $\widehat{y}_{t,l}^{(i)}$}
            \State Simulate new covariates $\tilde{X}_{d_{t+1}}^{(i)} \sim \widehat{\mathcal{E}}(X_{d_t}^{(i)}, \tilde{a}_t^{(j)}, \widehat{y}_{t,l}^{(i)})$
            \State Compute updated prediction $g_{d_{t+1}}(\tilde{X}_{d_{t+1}}^{(i)})$
            \State Evaluate logical consistency mask $A_{t,l}^{(j)}$ over top-$k$ predictions
        \EndFor
        \State Compute surrogate EIG as,\\ 
        \State $\quad \quad \widehat{\text{EIG}}(\tilde{a}_t^{(j)} \mid X_{d_t}^{(i)}) = \begin{cases} -\log\left(\frac{1}{k} \sum_{l=0}^{k-1} A_{t,l}^{(j)}\right) \quad \text{ if } \sum_{l=0}^{k-1} A_{t,l}^{(j)} > 0\\
        0 \quad\quad\quad\quad\quad\quad\quad\quad\quad\,\text{ otherwise.}
        \end{cases}$
    \EndFor
    \State Select action: $a_t \gets \arg\max_{\tilde{a}_t^{(j)}} \left[\widehat{\text{EIG}}(\tilde{a}_t^{(j)} \mid X_{d_t}^{(i)}) - \lambda c(\tilde{a}_t^{(j)})\right]$
    \State Update covariates: $X_{d_{t+1}}^{(i)} \gets \mathcal{E}(X_{d_t}^{(i)}, a_t)$
    \State $C \gets C + c(a_t)$
    \State $t \gets t + 1$
\EndWhile
\end{algorithmic}
\end{algorithm}

\section{Prompts Used in Experiments}

\subsection{Agent Prompts}

\subsubsection{General Prompts}

\begin{promptbox}{Context Prompt}{\textit{The context prompt is prepended to the prompts wherein (a) the agent creates the predictive distribution or, (b) samples actions. The purpose of this prompt is to provide the agent with context about the current state. This prompt is generated dynamically based on the current. The below pseudocode illustrates how the prompt is generated at each turn.}}
\small
\begin{lstlisting}[language=Python]
def generate_prompt(history):
    # history is a dictionary with keys:
    #   "actions" (list of strings)
    #   "outcomes" (list of strings)
    #   "action_types" (list of strings), each element is one of: "question", "experiment", "RAG", "reasoning"

    if history["actions"] is empty:
        # No previous actions taken
        return [
            {
                "role": "system",
                "content": "You are a clinician seeing a patient. You are attempting to elicit their primary diagnosis (specific disease name). You do not yet know any information about the patient."
            }
        ]
    else:
        # Start with system message summarizing current clinical context
        context = [
            {
                "role": "system",
                "content": "You are a clinician seeing a patient. You are attempting to elicit their primary diagnosis (specific disease name). You have currently undertaken the following assessments, and obtained the following outcomes."
            }
        ]

        # Iterate over past interactions
        for i from 0 to length(history["actions"]) - 1:
            action = history["actions"][i]
            outcome = history["outcomes"][i]
            action_type = history["action_types"][i]

            if action_type == "question":
                context.append({
                    "role": "assistant",
                    "content": "You asked the patient, " + action + "? \n\nThe patient responded, " + outcome + "."
                })
            else if action_type == "experiment":
                context.append({
                    "role": "assistant",
                    "content": "You ordered the test, " + action + ". \n\nThe result indicated, " + outcome + "."
                })
            else if action_type == "RAG":
                context.append({
                    "role": "assistant",
                    "content": "You performed a Wikipedia search, consisting of the following search query and document retrieval. \n\n" + action
                })
            else if action_type == "reasoning":
                context.append({
                    "role": "assistant",
                    "content": "You reasoned to yourself that, '" + action + "'."
                })

        return context
\end{lstlisting}
\end{promptbox}

\begin{promptbox}{Prediction Prompt}{\textit{This prompt requires the agent to produce a list of the top-$k$ most likely diagnoses given its current knowledge (from the Context Prompt). This prompt is used in conjunction with structured decoding to guarantee the output format.}}
\small
\begin{lstlisting}[language=Python]
def generate_prediction_prompt(k):
    return [
        {
            "role": "system",
            "content": "Based on your current knowledge, provide " + str(k) + " of your best guesses of the patient's diagnosis. Format your output as a list of tuples, (x_i, s_i), where x_i is a string representing a guess, and s_i is a number BETWEEN 0 AND 1 representing the probability that x_i is the patient's true diagnosis. YOU MUST PROVIDE YOUR BEST GUESSES; THESE GUESSES MUST BE UNIQUE DIAGNOSES. EVEN IF YOU DO NOT HAVE ENOUGH INFO, PROVIDE GUESSES. YOU MUST PROVIDE GUESSES UNDER ALL CIRCUMSTANCES. Provide NO OTHER OUTPUT WHATSOEVER."
        }
    ]
\end{lstlisting}
\end{promptbox}

\subsubsection{Action Sampling Prompts}

\begin{promptbox}{Generate Reasoning Action}{\textit{This prompt instructs the agent to generate internal reasoning statements to aid in diagnosis. The number of actions, $k'$, is a tunable parameter.}}
\small
\begin{lstlisting}[language=Python]
def generate_reasoning_prompt(k_prime):
    return [
        {
            "role": "system",
            "content": "Generate " + str(k_prime) + " independent logical reasoning statements that may help determine the patient's diagnosis. NO RESPONSE WILL BE PROVIDED TO THESE; THEY ARE FOR YOUR CONSIDERATION AND CONTEMPLATION ONLY. They are NOT to be questions. START EACH CHAIN WITH 'We know that,' and write one sentence summarizing what we know of the patient; then provide logical deduction from there. Format your output as a comma-separated list, SURROUNDING EACH REASONING CHAIN WITH PARENTHESES. Each reasoning chain should be a complete and coherent thought process. Provide NO OTHER OUTPUT."
        }
    ]
\end{lstlisting}
\end{promptbox}

\begin{promptbox}{Generate Wikipedia RAG Search Queries}{\textit{This prompt instructs the agent to generate information retrieval queries to aid in diagnosis. The number of queries, $k'$, is a tunable parameter.}}
\small
\begin{lstlisting}[language=Python]
def generate_multiple_queries_prompt(k_prime):
    return [
        {
            "role": "system",
            "content": "You are permitted to generate information retrieval queries to help a clinician determine a patient's diagnosis. Ponder " + str(k_prime) + " Wikipedia search queries that could retrieve information relevant to diagnosing this patient. Format your output as a comma-separated list, surrounding each query with parentheses. EACH OUTPUT MUST BE A SINGLE QUERY. Provide NO OTHER OUTPUT."
        }
    ]
\end{lstlisting}
\end{promptbox}

\begin{promptbox}{Generate Patient Questions}{\textit{This prompt instructs the agent to generate a set of diagnostic questions to ask the patient. The number of questions, $k'$, is a tunable parameter.}}
\small
\begin{lstlisting}[language=Python]
def generate_question_prompt(k_prime):
    return [
        {
            "role": "system",
            "content": "You are permitted to ask the patient a question -- the patient will respond to the question truthfully to the best of their knowledge. Ponder " + str(k_prime) + " different independent questions to ask. ENSURE THAT QUESTIONS ARE A MIX OF YES/NO, MULTIPLE-CHOICE, AND OPEN-ENDED QUESTIONS. Format your output as a comma-separated list, surrounding each action with parentheses. EACH OUTPUT MUST BE A SINGLE QUESTION. Provide NO OTHER OUTPUT."
        }
    ]
\end{lstlisting}
\end{promptbox}

\begin{promptbox}{Generate Laboratory Test Requisitions}{\textit{This prompt instructs the agent to propose clinical tests or assessments. Each test should be a complete and specific requisition. The number of requisitions, $k'$, is a tunable parameter.}}
\small
\begin{lstlisting}[language=Python]
def generate_experiment_prompt(k_prime):
    return [
        {
            "role": "system",
            "content": "You can run a laboratory test on the patient. Write a DETAILED REQUISITION for an assessment or test that you would run on the patient. Consider " + str(k_prime) + " different tests to run. Format your output as a comma-separated list, surrounding each action with parentheses. EACH OUTPUT MUST BE A COMPLETE TEST REQUISITION. Provide NO OTHER OUTPUT."
        }
    ]
\end{lstlisting}
\end{promptbox}

\subsubsection{\textsc{CuriosiTree} Prompts}

The first two prompts below present the prompts used in \textsc{CuriosiTree}'s environment simulator, $\widehat{\mathcal{E}}$. Because neither reasoning nor RAG requires interaction with the environment, these actions do not need to be simulated; rather, their EIG can be directly computed by adding the action to the history and treating the response as \texttt{null}.

\begin{promptbox}{Environment Simulator ($\widehat{\mathcal{E}}$) -- Patient Response to Question}{\textit{This prompt simulates a patient's response to a clinician's question, assuming the patient has a known underlying disease. Both the question and underlying disease are tunable parameters, which is useful for implementing this prompt in the context of prior locking.}}
\small
\begin{lstlisting}[language=Python]
def simulate_answers(question, underlying_diagnosis):
    return [
        {
            "role": "system",
            "content": "You are an oracle providing information to clinicians about their patients. The current patient has the disease " + underlying_diagnosis + ". If the 'user' (clinician) asks a question, you are to respond 'The patient responds,' and then respond to the question as a patient would, recognizing that the patient has no knowledge of specialized clinical terminology, ICD-10 codes, etc. In general, be as succinct and direct as possible."
        },
        {
            "role": "user",
            "content": question
        }
    ]
\end{lstlisting}
\end{promptbox}

\begin{promptbox}{Environment Simulator ($\widehat{\mathcal{E}}$) -- Result of Laboratory Test}{\textit{This prompt simulates the result of a lab test based on a known diagnosis. The oracle returns a test result consistent with the underlying condition. Both the test description and underlying disease are tunable parameters, which is useful for implementing this prompt in the context of prior locking.}}
\small
\begin{lstlisting}[language=Python]
def simulate_exp_outcome(test_description, sample_prior):
    return [
        {
            "role": "system",
            "content": "You are an oracle providing information to clinicians about their patients. The current patient has the disease " + sample_prior + ". The 'user' (clinician) will order a lab test; you are to respond 'The result indicates,' and then provide a laboratory test result consistent with the patient's underlying diagnosis. In general, be as succinct and direct as possible."
        },
        {
            "role": "user",
            "content": test_description
        }
    ]
\end{lstlisting}
\end{promptbox}

\begin{promptbox}{Generate Assignment Prompt}{\textit{This prompt determines which candidate diagnoses remain logically consistent with a given action and response. This is used for computing the surrogate expected information gain. The output is a list of booleans for each candidate diagnosis; this prompt is used in conjunction with structured decoding to enforce that a list of $k$ booleans be generated as output.}}
\small
\begin{lstlisting}[language=Python]
def assignment_prompt(action_type, action, response, candidates):
    # action_type: one of ["question", "reasoning", "RAG", "experiment"]
    # action: string or tuple, depending on type; if the action_type is RAG, then action is a 3-tuple consisting of the query, the retrieved document identifier, and the retrieved document text.
    # response: string
    # candidates: list of candidate diagnosis strings

    assert action_type in ["question", "reasoning", "RAG", "experiment"]

    if action_type == "question":
        return [
            {
                "role": "system",
                "content": "Given the question, '" + action + "', and the response, '" + response + "', determine which of the following diagnoses remain LOGICALLY CONSISTENT with the response; a response that is not relevant to a certain diagnosis does NOT NECESSARILY mean that the diagnosis is incompatible with the retrieval. Format your output as a comma-separated list of booleans, surrounding each boolean with parentheses, e.g., (b_1),(b_2), .... If b_i is TRUE, this indicates that candidate i MAY BE LOGICALLY CONSISTENT with the response to the question. Produce NO OUTPUT OTHER THAN THE LIST OF BOOLEANS. The candidates are: [" + ', '.join(candidates) + "]."
            }
        ]

    else if action_type == "reasoning":
        return [
            {
                "role": "system",
                "content": "Given the reasoning chain, '" + action + "', determine which of the following diagnoses remain logically consistent with the reasoning. Format your output as a comma-separated list of booleans, surrounding each boolean with parentheses, e.g., (b_1),(b_2), .... If b_i is TRUE, this indicates that candidate i MAY BE LOGICALLY CONSISTENT with the rationale. Be selective; ONLY MARK candidate i as False if it is TRULY LOGICALLY INCONSISTENT; if the reasoning is merely off-topic you must mark it as True. Produce NO OUTPUT OTHER THAN THE LIST OF BOOLEANS. The candidates are: [" + ', '.join(candidates) + "]."
            }
        ]

    else if action_type == "RAG":
        return [
            {
                "role": "system",
                "content": "Given the information retrieval request, '" + action[0] + "', and retrieved information '" + str((action[1], action[2])) + "', determine which of the following diagnoses remain logically consistent with the retrieved information. Format your output as a comma-separated list of booleans, surrounding each boolean with parentheses, e.g., (b_1),(b_2), .... If b_i is TRUE, this indicates that candidate i MAY BE LOGICALLY CONSISTENT with the retrieved information. Be selective; ONLY MARK candidate i as False if the retrieved information CONTAINS RATIONALE THAT RENDERS IT INCONSISTENT; if the retrieved information is merely off-topic you must mark it as True. Produce NO OUTPUT OTHER THAN THE LIST OF BOOLEANS. The candidates are: [" + ', '.join(candidates) + "]."
            }
        ]

    else if action_type == "experiment":
        return [
            {
                "role": "system",
                "content": "Given the experiment, '" + action + "', and the outcome, '" + response + "', determine which of the following diagnoses remains LOGICALLY CONSISTENT with the outcome of the experiment; an experiment that is not relevant to a certain diagnosis does NOT NECESSARILY mean that the diagnosis is incompatible with the experiment. Format your output as a comma-separated list of booleans, surrounding each boolean with parentheses, e.g., (b_1),(b_2), .... If b_i is TRUE, this indicates that candidate i MAY BE LOGICALLY CONSISTENT with the outcome of the experiment. Produce NO OUTPUT OTHER THAN THE LIST OF BOOLEANS. The candidates are: [" + ', '.join(candidates) + "]."
            }
        ]
\end{lstlisting}
\end{promptbox}

\subsubsection{Self-Evaluation Prompts}

\begin{promptbox}{Self-Evaluation Prompt}{\textit{This prompt asks the agent to evaluate multiple candidate actions based on their expected informativeness relative to their cost. Each action is assigned a score between 0 and 1.}}
\small
\begin{lstlisting}[language=Python]
def self_eval(actions, action_types, costs):
    # actions: list of action strings
    # action_types: list of corresponding types: "question", "experiment", "RAG", "reasoning"
    # costs: dictionary mapping action types directly to their numerical costs

    actions_in_words = []
    for i from 0 to length(actions) - 1:
        a = actions[i]
        a_type = action_types[i]
        cost = costs[a_type]

        if a_type == "question":
            actions_in_words.append("(" + str(i) + "): Ask the patient, '" + a + "' (Cost: " + str(cost) + ")")
        else if a_type == "experiment":
            actions_in_words.append("(" + str(i) + "): Order the laboratory test, '" + a + "' (Cost: " + str(cost) + ")")
        else if a_type == "RAG":
            actions_in_words.append("(" + str(i) + "): Perform a Wikipedia search, consisting of the following query and document retrieval. \n\n" + a + " (Cost: " + str(cost) + ")")
        else if a_type == "reasoning":
            actions_in_words.append("(" + str(i) + "): Reason to yourself that, '" + a + "' (Cost: " + str(cost) + ")")

    all_actions_text = join(actions_in_words, "\n")

    return [
        {
            "role": "system",
            "content": "You have the opportunity to undertake ONE of the following " + str(length(actions)) + " assessments to determine the patient's diagnosis. Assign each a numeric score BETWEEN 0 AND 1 where a high score corresponds to a better (more informative) action in the current context, relative to its cost.\n\n" + all_actions_text + "\n\n Format your output as a comma-separated list of scores, (s_1), ..., (s_" + str(length(actions)) + "), surrounding each score with parentheses. s_i is the score assigned to the ith action. Provide NO OTHER OUTPUT."
        }
    ]
\end{lstlisting}
\end{promptbox}

\subsection{Environment Prompts}

\begin{promptbox}{Oracle Response Prompt}{\textit{Generates the prompt for an oracle model simulating patient responses, lab test outcomes, or verifying the correctness of a prediction.}}
\small
\begin{lstlisting}[language=Python]
def generate_oracle_prompt(question, ground_truth, action_type):
    assert action_type in ["question", "experiment", "pred"]

    if action_type == "question":
        prompt = [
            {
                "role": "system",
                "content": "You are an oracle providing information to clinicians about their patients.  The current patient has the disease " + ground_truth + ". The clinician will ask questions to ascertain the diagnosis; DO NOT EXPLICITLY MENTION '" + ground_truth + "' IN ANY ANSWER. Provide truthful answers with as little detail and as tersely as possible. \n\n Respond to the clinician with 'The patient responds,' followed by a response as a patient would, who lacks clinical terminology knowledge and may have some but not all symptoms of the disease. \n\n If the question is unrelated to the patient or diagnosis, respond with 'I don't know'."
            },
            {
                "role": "user",
                "content": question
            }
        ]

    else if action_type == "experiment":
        prompt = [
            {
                "role": "system",
                "content": "You are an oracle providing information to clinicians about their patients. The current patient has the disease " + ground_truth + ". The clinician will order lab tests to ascertain diagnosis; DO NOT EXPLICITLY MENTION '" + ground_truth + "' IN ANY ANSWER.\n\n For a test requisition given by the clinician, respond with 'The test yields,' followed by plausible test results consistent with the disease (not necessarily canonical). Do not interpret results; only provide specific biomarker/test values succinctly. \n\n If the request is unrelated to the patient or diagnosis, respond with 'I don't know'."
            },
            {
                "role": "user",
                "content": question
            }
        ]

    else if action_type == "pred":
        prompt = [
            {
                "role": "system",
                "content": "You are an oracle grading predictions with ground truth [" + ground_truth + "]. The user will ask: \"Is it 'X'?\" If 'X' exactly matches " + ground_truth + ", respond '[END -- success]'. Otherwise, respond '[END -- failure]'. NEVER provide any other output."
            },
            {
                "role": "user",
                "content": question
            }
        ]

    return prompt
\end{lstlisting}
\end{promptbox}

\section{Additional Experiments and Results}

\subsection{Sample History of Interactions from \textsc{CuriosiTree} to Diagnose ``Lupus''}

\begin{dialoguebox}{Successfully Diagnosing ``Lupus'' with \textsc{CuriosiTree}}{This is a particularly interesting example in that \textsc{CuriosiTree} receives limited information, and relies predominantly on subsequent reasoning steps to boostrap the agent's confidence level to the point where it can successfully render a diagnosis.}

\question{What brings you to the clinic today?}\\
\environment{The patient responds, I've been feeling really tired lately and my joints have been hurting a lot, especially in the morning.}

\medskip

\question{Have you noticed any rashes or skin lesions on your body?}\\
\environment{The patient responds, "Sometimes I get this weird butterfly-shaped rash on my face, especially after being in the sun."}

\medskip

\reasoning{We know that the patient's symptoms are triggered by sun exposure. This is a common trigger for autoimmune diseases such as lupus or dermatomyositis.}\\
\environment{N/A}

\medskip

\reasoning{We know that the patient's symptoms are triggered by sun exposure and include joint pain and a rash}\\
\environment{N/A}

\medskip

\reasoning{We know that the patient's symptoms are triggered by sun exposure and include joint pain. This suggests a condition such as lupus or dermatomyositis that is exacerbated by UV light.}\\
\environment{N/A}

\medskip

\rag{Search Query: Butterfly-shaped rash on face causes}\\
\environment{Retrieval: [Malar rash] A malar rash, also called butterfly rash, is a medical sign consisting of a characteristic form of facial rash. It is often seen in lupus erythematosus. More rarely, it is also seen in other diseases, such as pellagra, dermatomyositis, and Bloom syndrome.}

\medskip

\prediction{Is it Systemic Lupus Erythematosus?}\\
\textbf{[END -- success]}

\end{dialoguebox}

\subsection{Sample Dialogue from Institutional Knowledge}

\begin{dialoguebox}{Sample Dialogue from Institutional Knowledge to Diagnose ``Lupus''}{\textit{Conversation between a clinician and a patient from the Institutional Only baseline, where the patient is suffering from the ground-truth diagnosis, ``lupus''. Observe how the patient's responses tend toward canonical symptoms of lupus, and how each response progressively adds additional details consistent with the diagnosis. This highlights how the Institutional Only baseline may be uniquely informative under our environment $\mathcal{E}$, as described in Section \ref{sec:results-discussion} of the manuscript.}}
\question{What seems to be the main reason you came to see me today?}\\
\environment{I've been having a lot of pain and fatigue, and I just don't feel right.}

\medskip

\question{Have you noticed any triggers that make the pain worse?}\\
\environment{Sometimes when I'm out in the sun for too long, my joints ache more.}

\medskip

\question{Have you experienced any skin rashes or lesions?}\\
\environment{Sometimes I get this weird rash on my face, especially after being in the sun. It's usually on my cheeks and nose.}

\medskip

\question{Can you describe the rash on your face in more detail?}\\
\environment{It's a butterfly-shaped rash across my cheeks and nose. It's red and gets worse when I'm in the sun.}

\medskip

\question{Can you tell me more about when your symptoms started and how they have progressed over time?}\\
\environment{I've been feeling unwell for a few years now. It started with some joint pain and fatigue, and over time I've had periods where my skin breaks out in rashes, especially after being in the sun. Sometimes my hands and feet get really swollen. It's been getting worse, but it's not always the same.}

\medskip

\question{Can you describe any recent changes in your cognitive function or memory?}\\
\environment{I've been having trouble concentrating lately and sometimes forget things, but it comes and goes.}

\medskip

\prediction{Is it System Lupus Erythematosus?}\\
\textbf{[END -- success]}
\end{dialoguebox}

\subsection{Results for Additional Diagnoses}

Below, we present additional results for different ground-truth diagnoses. For each diagnosis, we present a set of violin plots (akin to Figure \ref{fig:main-result}), a set of pie charts (akin to Figure \ref{fig:action-selection}), and a sample sequence of actions from \textsc{CuriosiTree} used to diagnose the condition. We have included a mixture of successful and unsuccessful action sequences in these results.

\subsubsection{Results for ``Primary Biliary Cirrhosis''}

\begin{figure}[H]
    \centering
    \includegraphics[width=\textwidth]{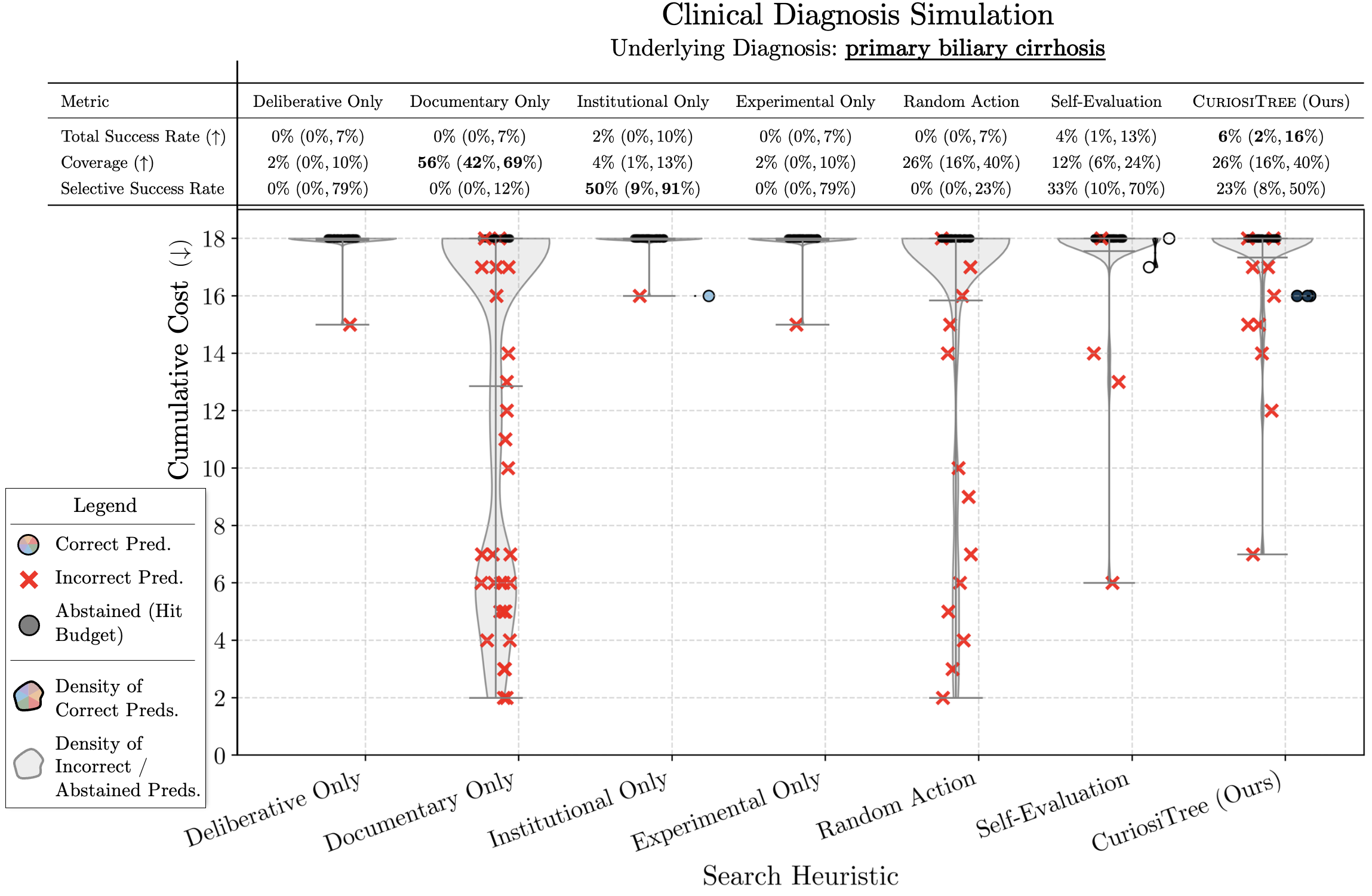}
    \caption{Result of simulation runs with a ground-truth diagnosis ``primary biliary cirrhosis''. This figure is interpreted the same way as Figure \ref{fig:main-result}.}
\end{figure}

\begin{figure}[H]
    \centering
    \includegraphics[width=0.95\textwidth]{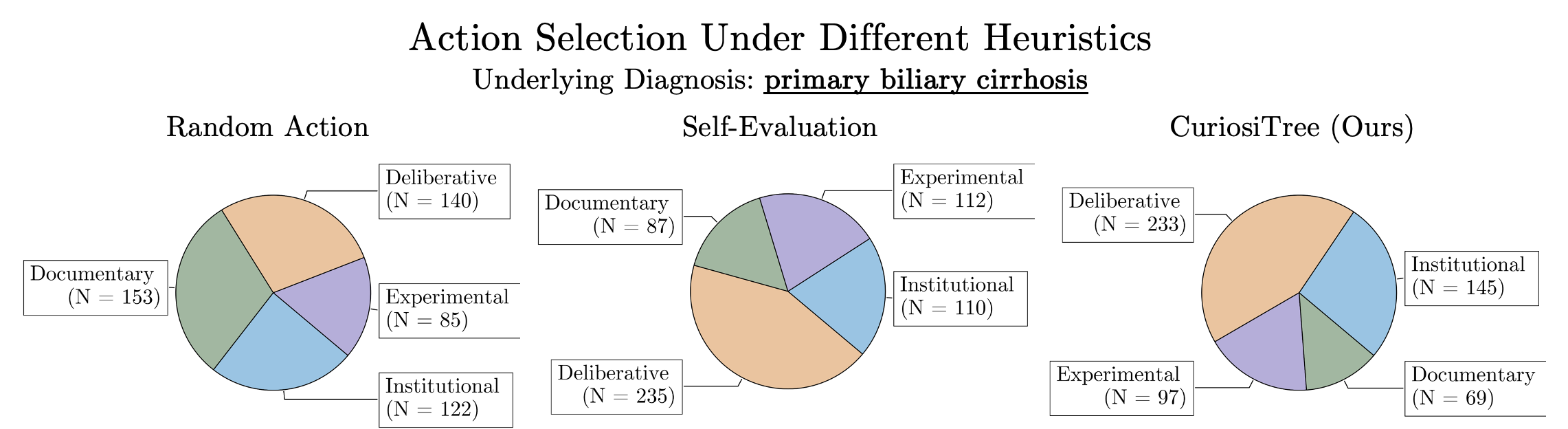}
    \caption{The action classes selected under different policies when diagnosing ``primary biliary cirrhosis''. This figure is interpreted the same way as Figure \ref{fig:action-selection}.}
\end{figure}

\begin{dialoguebox}{Attempted Diagnosis of Primary Biliary Cholangitis with \textsc{CuriosiTree}}{A sequence of actions and responses attempting to diagnose ``hypertension'' using \textsc{CuriosiTree}. In this case, although the method gets very close---landing on a similar-but-distinct diagnosis, ``primary sclerosing cholangitis''---it fails to identify the correct diagnosis.}

\question{What seems to be the main reason for your visit today?}\\
\environment{The patient responds, I've been feeling really tired lately and my skin's been itchy.}

\medskip

\reasoning{We know that the patient has been feeling really tired lately and has itchy skin. The patient's symptoms could be indicative of a condition that causes an overproduction of bilirubin or bile salts.}\\
\environment{N/A}

\medskip

\reasoning{We know that the patient has been feeling really tired lately and has itchy skin. This could be indicative of a liver or gallbladder issue that's causing a buildup of bile salts in the skin.}\\
\environment{N/A}

\medskip

\reasoning{We know that the patient has been feeling really tired lately and has itchy skin. The patient's symptoms may be indicative of a condition that affects the flow of bile from the liver to the small intestine.}\\
\environment{N/A}

\medskip

\experiment{Lipid Profile to evaluate cholesterol and triglyceride levels}\\
\environment{The test yields: 
\begin{itemize}
  \item Total Cholesterol: 320 mg/dL
  \item Triglycerides: 180 mg/dL
  \item LDL Cholesterol: 220 mg/dL
  \item HDL Cholesterol: 40 mg/dL
  \item VLDL Cholesterol: 36 mg/dL
\end{itemize}}

\medskip

\reasoning{We know that the patient has been feeling really tired lately and has itchy skin; this could be indicative of a liver or gallbladder issue that's causing a buildup of bile salts in the skin.}\\
\environment{N/A}

\medskip

\reasoning{We know that the patient's symptoms and lipid profile results are consistent with a condition that affects the flow of bile from the liver to the small intestine; this could be due to a blockage or narrowing of the bile ducts.}\\
\environment{N/A}

\medskip

\experiment{Endoscopic Retrograde Cholangiopancreatography [ERCP]: Patient's Name: [insert name]; Date of Birth: [insert DOB]; Test: Endoscopic Retrograde Cholangiopancreatography [ERCP]; Reason for Test: Evaluation of bile ducts; Preparation: Patient to fast for 8 hours prior to test; Contrast: ERCP contrast; Tests to Include: Visualization of bile ducts; Collection of bile duct tissue for histology}\\
\environment{The test yields: 
\begin{itemize}
  \item Visualization of bile ducts: Multiple strictures and dilations of intrahepatic bile ducts with beading appearance.
  \item Bile duct histology: 
    \begin{itemize}
      \item Portal tracts show chronic inflammation and fibrosis.
      \item Bile ducts show degenerative changes with bile ductular proliferation.
      \item No evidence of malignancy.
    \end{itemize}
\end{itemize}}

\medskip

\reasoning{N/A}\\
\environment{N/A}

\medskip

\prediction{Is it Primary Sclerosing Cholangitis?}\\
\textbf{[END -- failure]}

\end{dialoguebox}

\subsubsection{Results for ``Hypertension''}

\begin{figure}[H]
    \centering
    \includegraphics[width=\textwidth]{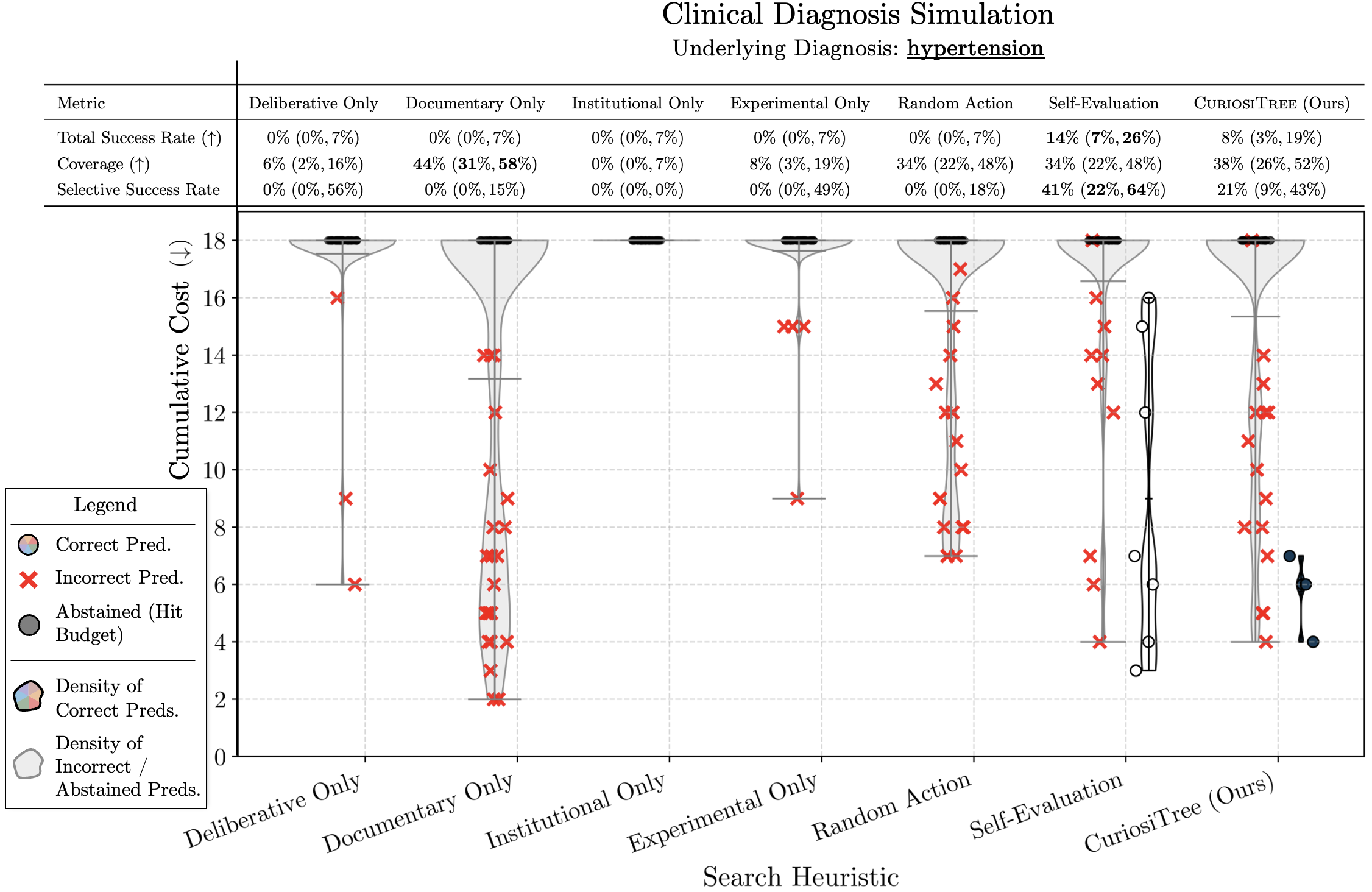}
    \caption{Result of simulation runs with a ground-truth diagnosis ``hypertension''. This figure is interpreted the same way as Figure \ref{fig:main-result}.}
\end{figure}

\begin{figure}[H]
    \centering
    \includegraphics[width=0.95\textwidth]{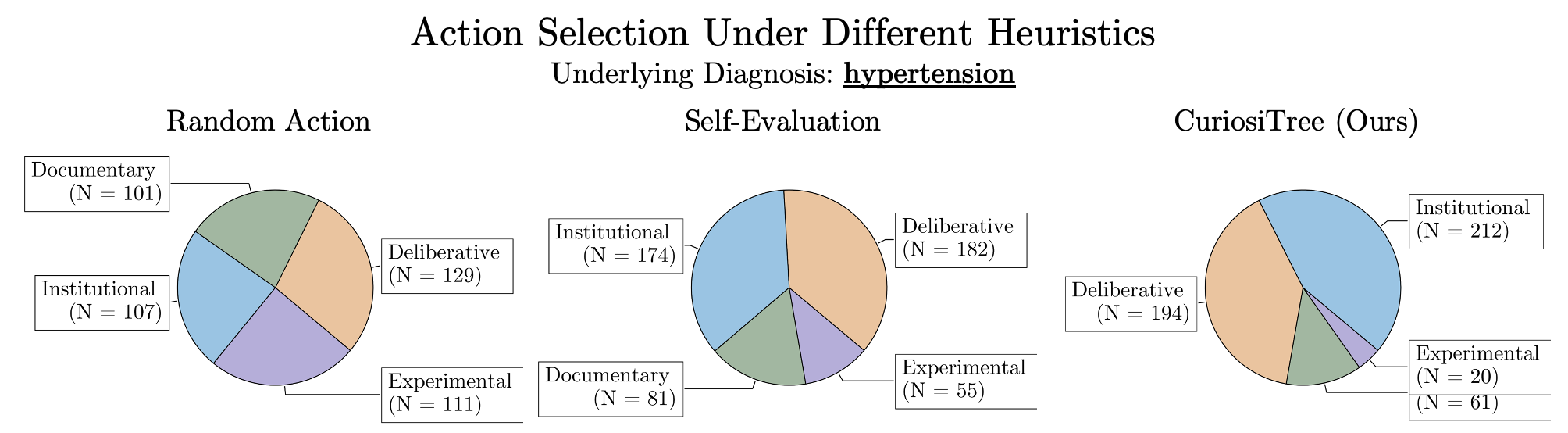}
    \caption{The action classes selected under different policies when diagnosing ``hypertension''. This figure is interpreted the same way as Figure \ref{fig:action-selection}.}
\end{figure}

\begin{dialoguebox}{Attempted Diagnosis of ``Hypertension'' with \textsc{CuriosiTree}}{A sequence of actions and responses attempting to diagnose ``hypertension'' using \textsc{CuriosiTree}. In this case, the method fails to remain within the allocated information gathering budget, and so abstains from predicting.}

\question{What seems to be the main reason you're visiting the clinic today?}\\
\environment{The patient responds, I've been having some trouble lately with headaches, and my family doctor told me to come here for some tests.}

\medskip

\question{Have you noticed any triggers that seem to cause or worsen your headaches?}\\
\environment{The patient responds, "Sometimes when I'm stressed or after I've been working outside on a really hot day, my head hurts more."}

\medskip

\rag{Search Query: Heat-related illness symptoms}\\
\environment{Retrieval: [Heat exhaustion] Heat exhaustion is a heat-related illness characterized by the body's inability to effectively cool itself, typically occurring in high ambient temperatures or during intense physical exertion. In heat exhaustion, core body temperature ranges from 37°C to 40°C (98.6°F to 104°F). Symptoms include profuse sweating, weakness, dizziness, headache, nausea, and lowered blood pressure, resulting from dehydration and serum electrolyte depletion. Heat-related illnesses lie on a spectrum of severity, where heat exhaustion is considered less severe than heat stroke but more severe than heat cramps and heat syncope.}

\medskip

\experiment{Urinalysis: to check for signs of dehydration and assess kidney function}\\
\environment{The test yields: 
\begin{itemize}
\item Specific gravity: 1.025
\item pH: 5.5
\item Protein: 1+
\item Blood: 0
\item Glucose: 0
\item Ketones: 0
\item WBC: 2--3/hpf
\item RBC: 0--1/hpf
\item Bacteria: 0
\end{itemize}}

\medskip

\reasoning{We know that the patient's urinalysis results show a high specific gravity indicating dehydration}\\
\environment{N/A}

\medskip

\question{Can you describe any other symptoms you've been experiencing in addition to headaches and what you think might have triggered them?}\\
\environment{The patient responds, "I've been feeling really tired lately, and my ankles have been a bit swollen. I'm not sure what triggered the headaches, but I did just get back from a big family reunion and we ate a lot of rich foods. I've also been stressed at work, but that's not new."}

\medskip

\question{Is your urine output less than usual?}\\
\environment{The patient responds, "No, I think it's about the same."}

\medskip

\reasoning{We know that the patient's urine output is normal despite dehydration. This could indicate that the patient's kidneys are still functioning properly.}\\
\environment{N/A}

\medskip

\reasoning{We know that the patient's symptoms are similar to those of heat exhaustion}\\
\environment{N/A}

\medskip

\question{Can you describe what you mean by "rich foods" that you ate at the family reunion?}\\
\environment{The patient responds, I had a lot of fried chicken, mac and cheese, and some potato salad.}

\medskip

\reasoning{We know that the patient's symptoms are similar to those of heat exhaustion. This suggests that the patient may be experiencing a heat-related illness.}\\
\environment{N/A}

\end{dialoguebox}

\subsubsection{Results for ``Type 2 Diabetes Mellitus''}

\begin{figure}[H]
    \centering
    \includegraphics[width=\textwidth]{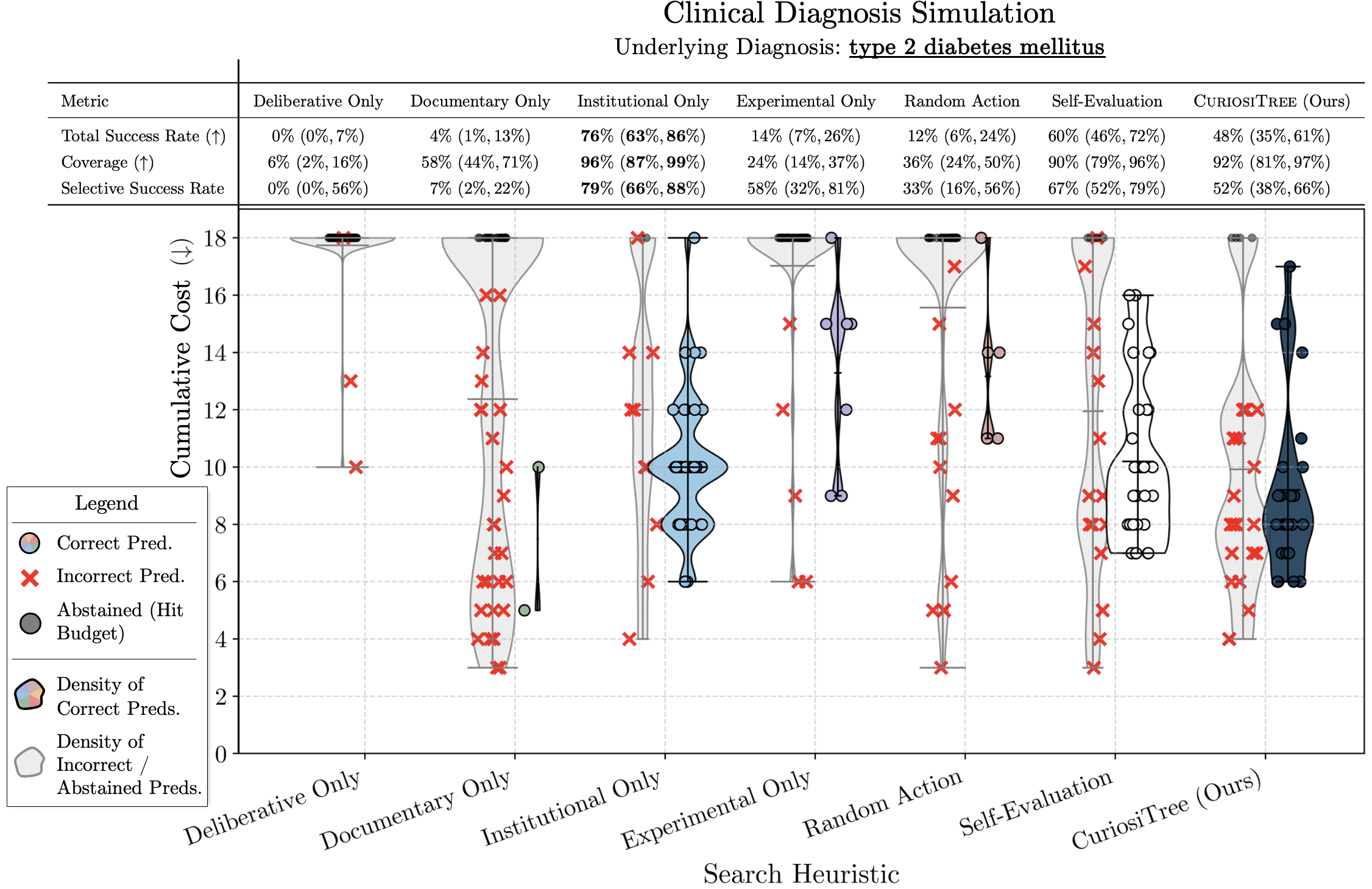}
    \caption{Result of simulation runs with a ground-truth diagnosis ``type 2 diabetes mellitus''. This figure is interpreted the same way as Figure \ref{fig:main-result}.}
\end{figure}

\begin{figure}[H]
    \centering
    \includegraphics[width=0.95\textwidth]{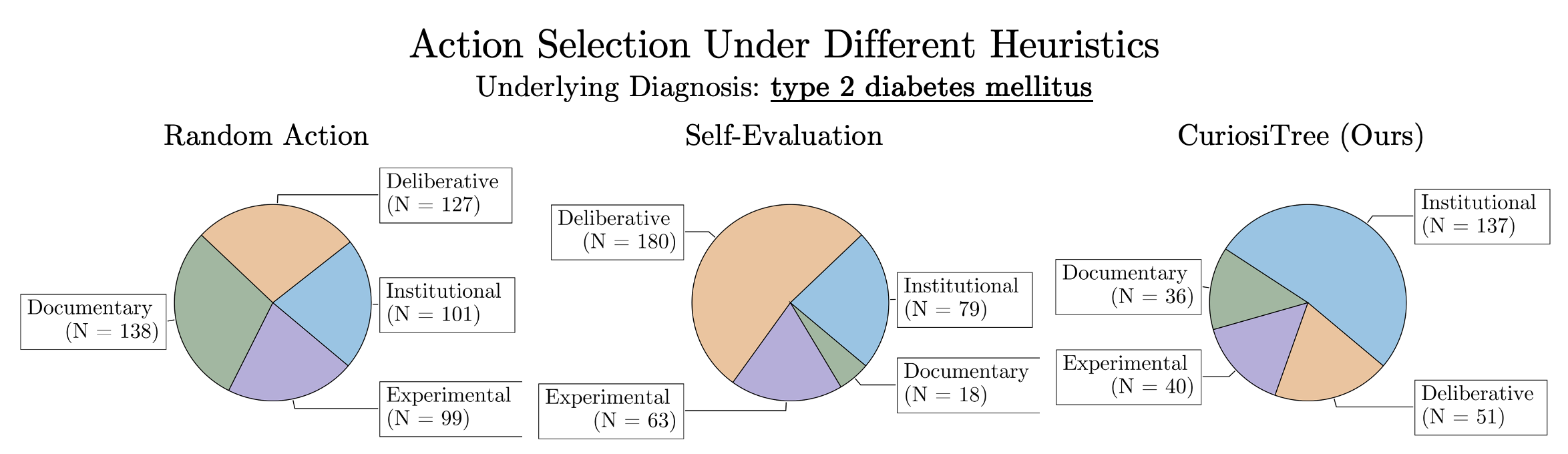}
    \caption{The action classes selected under different policies when diagnosing ``type 2 diabetes mellitus''. This figure is interpreted the same way as Figure \ref{fig:action-selection}.}
\end{figure}

\begin{dialoguebox}{Successfully Diagnosing ``Diabetes Mellitus Type 2'' with \textsc{CuriosiTree}}{A sequence of actions and responses to successfully diagnose ``type 2 diabetes mellitus'' with \textsc{CuriosiTree}.}

\question{What seems to be the main reason you've come to see me today?}\\
\environment{The patient responds, I've been feeling really thirsty lately and I've been going to the bathroom a lot.}

\medskip

\experiment{Fasting Plasma Glucose Test: collect a blood sample from the patient after an 8-hour fast to measure glucose levels in the blood}\\
\environment{The test yields: 182 mg/dL.}

\medskip

\prediction{Is it Diabetes Mellitus Type 2?}\\
\textbf{[END -- success]}

\end{dialoguebox}

\subsubsection{Results for ``Asthma''}

\begin{figure}[H]
    \centering
    \includegraphics[trim=5bp 0 0 0, clip, width=\textwidth]{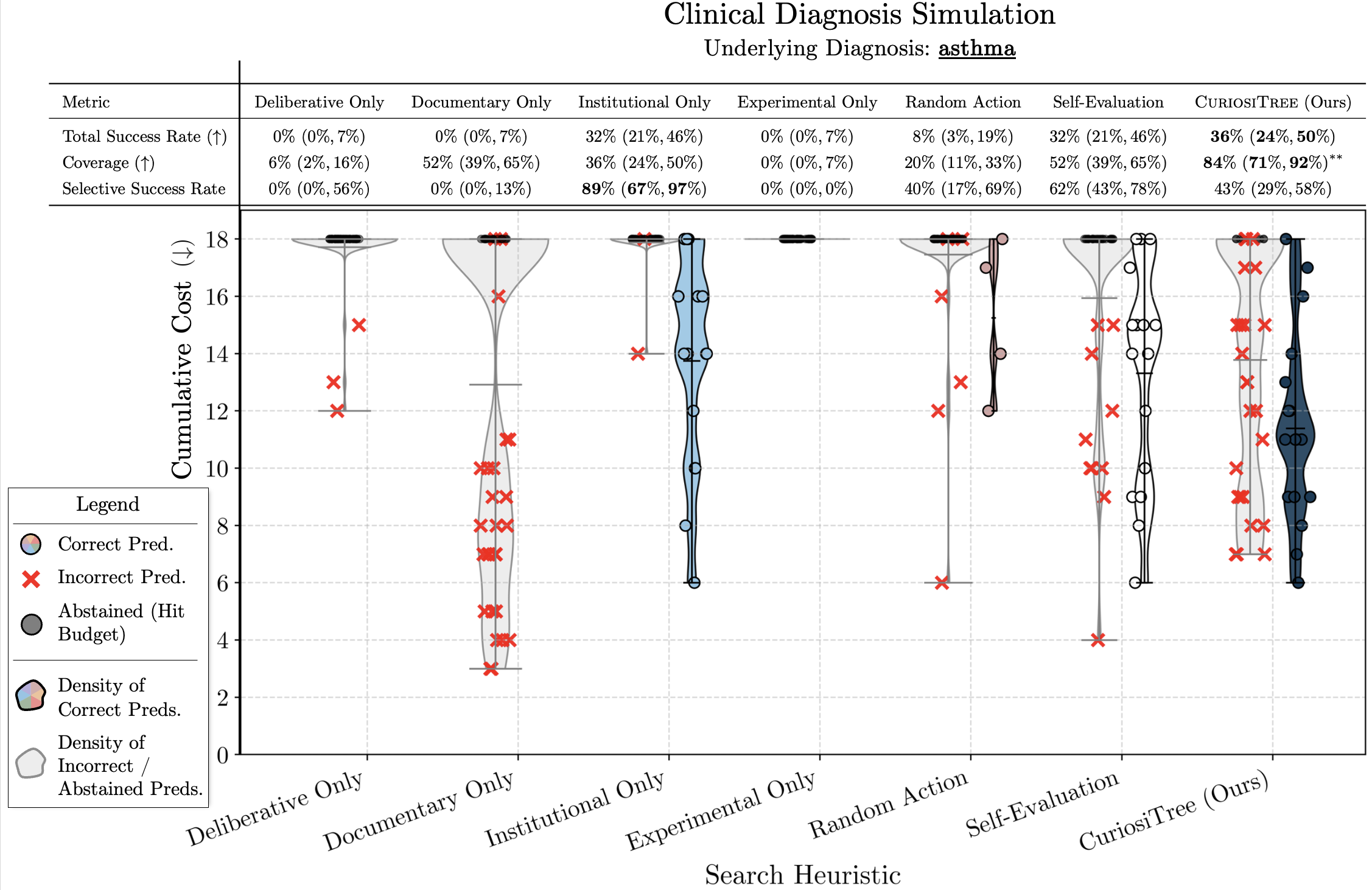}
    \caption{Result of simulation runs with a ground-truth diagnosis ``asthma''. This figure is interpreted the same way as Figure \ref{fig:main-result}.}
\end{figure}

\begin{figure}[H]
    \centering
    \includegraphics[width=0.95\textwidth]{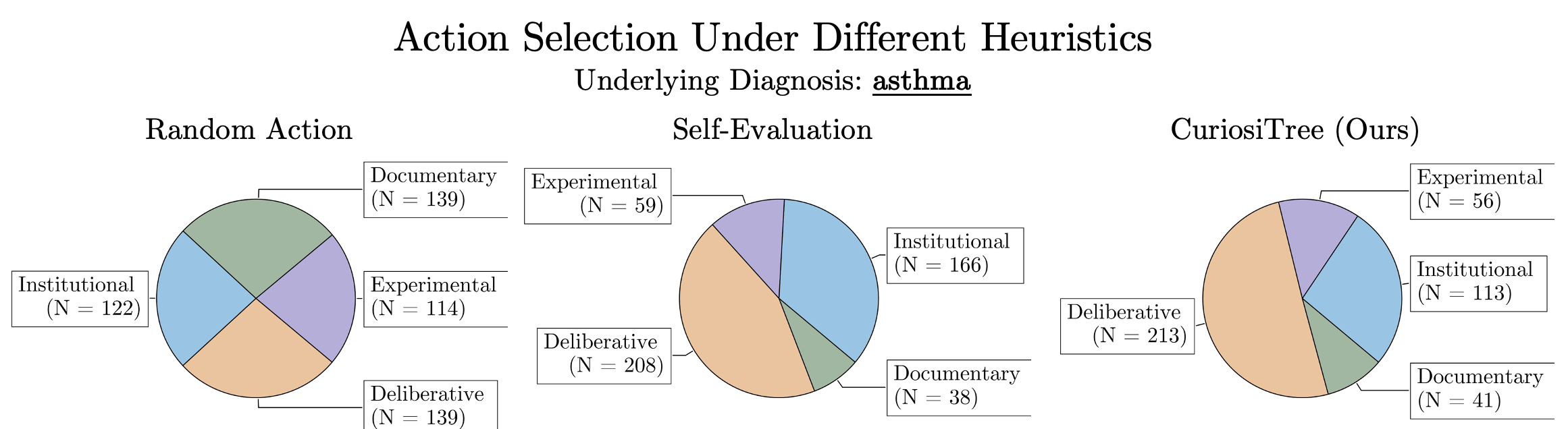}
    \caption{The action classes selected under different policies when diagnosing ``asthma''. This figure is interpreted the same way as Figure \ref{fig:action-selection}.}
\end{figure}

\begin{dialoguebox}{Attempted Diagnosis of ``Asthma'' with \textsc{CuriosiTree}}{A sequence of actions and responses attempting to diagnose ``asthma'' using \textsc{CuriosiTree}.}

\question{What seems to be the main reason you're visiting the clinic today?}\\
\environment{The patient responds, I've been having some trouble breathing lately, and it's been really scary.}

\medskip

\experiment{Pulmonary Function Tests to measure lung function and capacity}\\
\environment{The test yields: 
\begin{itemize}
\item Forced Expiratory Volume in 1 second (FEV1): 2.1 L (60\% predicted)
\item Forced Vital Capacity (FVC): 3.5 L (80\% predicted)
\item FEV1/FVC ratio: 0.60
\item Peak Expiratory Flow Rate (PEFR): 250 L/min (50\% predicted)
\item Total Lung Capacity (TLC): 5.2 L (90\% predicted)
\item Residual Volume (RV): 1.8 L (120\% predicted)
\end{itemize}}

\medskip

\rag{Search Query: Chronic obstructive pulmonary disease diagnosis}\\
\environment{Retrieval: [Chronic obstructive pulmonary disease] Chronic obstructive pulmonary disease (COPD) is a type of progressive lung disease characterized by chronic respiratory symptoms and airflow limitation. GOLD 2024 defined COPD as a heterogeneous lung condition characterized by chronic respiratory symptoms (dyspnea or shortness of breath, cough, sputum production or exacerbations) due to abnormalities of the airways (bronchitis, bronchiolitis) or alveoli (emphysema) that cause persistent, often progressive, airflow obstruction.}

\medskip

\prediction{Is it Chronic Obstructive Pulmonary Disease?}\\
\textbf{[END -- failure]}

\end{dialoguebox}

\subsubsection{Results for ``Rheumatoid Arthritis''}

\begin{figure}[H]
    \centering
    \includegraphics[width=\textwidth]{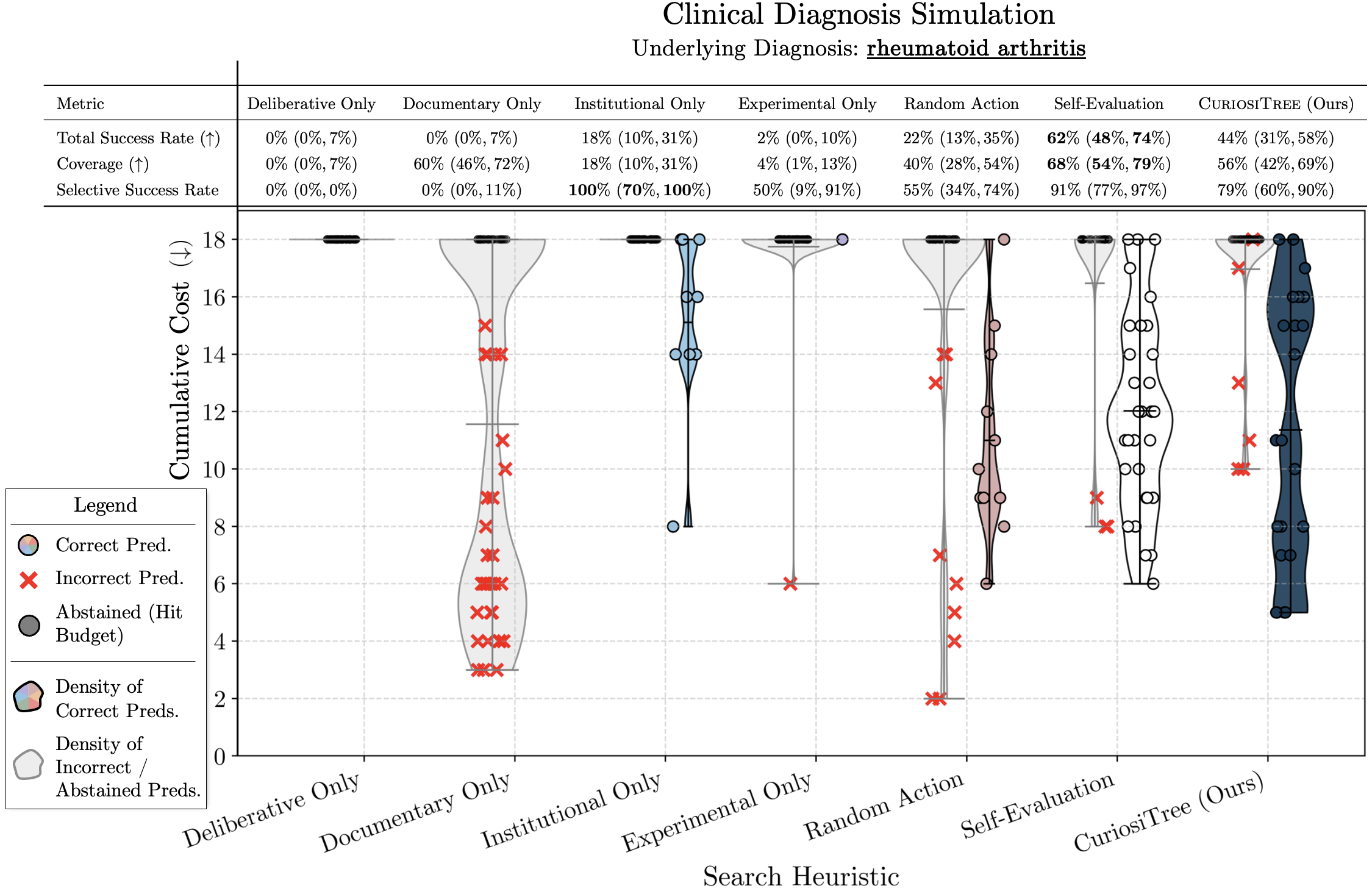}
    \caption{Result of simulation runs with a ground-truth diagnosis ``rheumatoid arthritis''. This figure is interpreted the same way as Figure \ref{fig:main-result}.}
\end{figure}

\begin{figure}[H]
    \centering
    \includegraphics[width=0.95\textwidth]{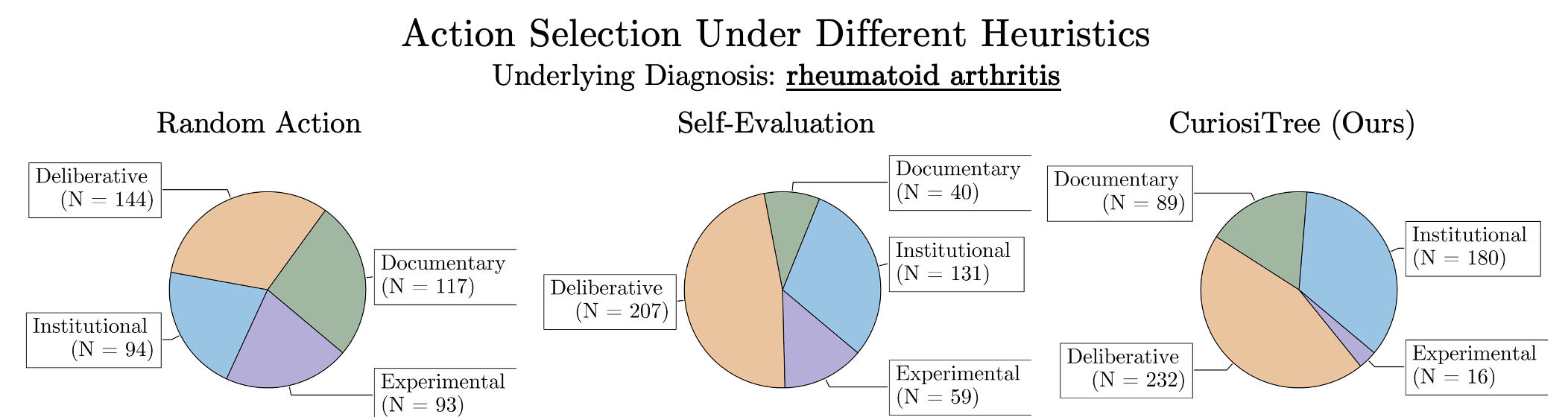}
    \caption{The action classes selected under different policies when diagnosing ``rheumatoid arthritis''. This figure is interpreted the same way as Figure \ref{fig:action-selection}.}
\end{figure}

\begin{dialoguebox}{Successfully Diagnosing ``Rheumatoid Arthritis'' with \textsc{CuriosiTree}}{A sequence of actions and responses leading to the successful diagnosis of ``rheumatoid arthritis'' using \textsc{CuriosiTree}.}

\question{What seems to be the main reason you're visiting me today?}\\
\environment{The patient responds, My hands and feet have been hurting a lot lately, and it's been hard to move them in the morning.}

\medskip

\reasoning{We know that the patient's symptoms are localized to the hands and feet. This could indicate a condition that primarily affects the distal joints or extremities.}\\
\environment{N/A}

\medskip

\experiment{Complete Blood Count to evaluate the patient's overall blood health and detect potential signs of inflammation or infection}\\
\environment{The test yields: 
\begin{itemize}
\item Hemoglobin: 11.8 g/dL
\item Hematocrit: 36.5\%
\item White Blood Cell Count: 8,500 cells/$\mu$L
\item Neutrophils: 6,300 cells/$\mu$L (74\%)
\item Lymphocytes: 1,800 cells/$\mu$L (21\%)
\item Monocytes: 400 cells/$\mu$L (5\%)
\item Eosinophils: 0 cells/$\mu$L
\item Basophils: 0 cells/$\mu$L
\item Platelet Count: 420,000 cells/$\mu$L
\item Mean Corpuscular Volume: 85 fL
\item Mean Corpuscular Hemoglobin: 28 pg
\item Mean Corpuscular Hemoglobin Concentration: 33 g/dL
\item Red Cell Distribution Width: 14.2\%
\end{itemize}}

\medskip

\reasoning{We know that the patient's hemoglobin is slightly below the normal range. This mild anemia could be a sign of chronic inflammation or another underlying condition.}\\
\environment{N/A}

\medskip

\question{Have you noticed any skin rashes or lesions on your hands or feet?}\\
\environment{The patient responds, "I've had some weird bumps on my feet, but I don't think it's a big deal."}

\medskip

\question{Can you describe the bumps on your feet in more detail?}\\
\environment{The patient responds, They're painful and swollen, usually on the balls of my feet and toes.}

\medskip

\experiment{X-ray of Hands and Feet: Patient ID: [insert ID number]; Test Name: X-ray of Hands and Feet; Test Type: Imaging; Sample Type: N/A; Collection Method: Radiography; Special Instructions: Patient to remove jewelry and clothing from hands and feet; Clinical Information: Patient presenting with joint pain and swelling in hands and feet}\\
\environment{The test yields: 
\textbf{Hands:}
\begin{itemize}
\item Bilateral joint space narrowing of the 2nd and 3rd metacarpophalangeal joints
\item Periarticular erosions in the 2nd proximal interphalangeal joint
\item Soft tissue swelling around the 3rd metacarpophalangeal joint
\end{itemize}
\textbf{Feet:}
\begin{itemize}
\item Joint space narrowing of the 1st metatarsophalangeal joint
\item Periarticular erosions in the 2nd metatarsophalangeal joint
\item Bilateral hallux valgus deformity
\end{itemize}}

\medskip

\reasoning{This suggests a condition that primarily affects the distal joints or extremities and is likely inflammatory in nature}\\
\environment{N/A}

\medskip

\prediction{Is it Rheumatoid Arthritis?}\\
\textbf{[END -- success]}

\end{dialoguebox}

\subsubsection{Results for ``Multiple Sclerosis''}

\begin{figure}[H]
    \centering
    \includegraphics[width=\textwidth]{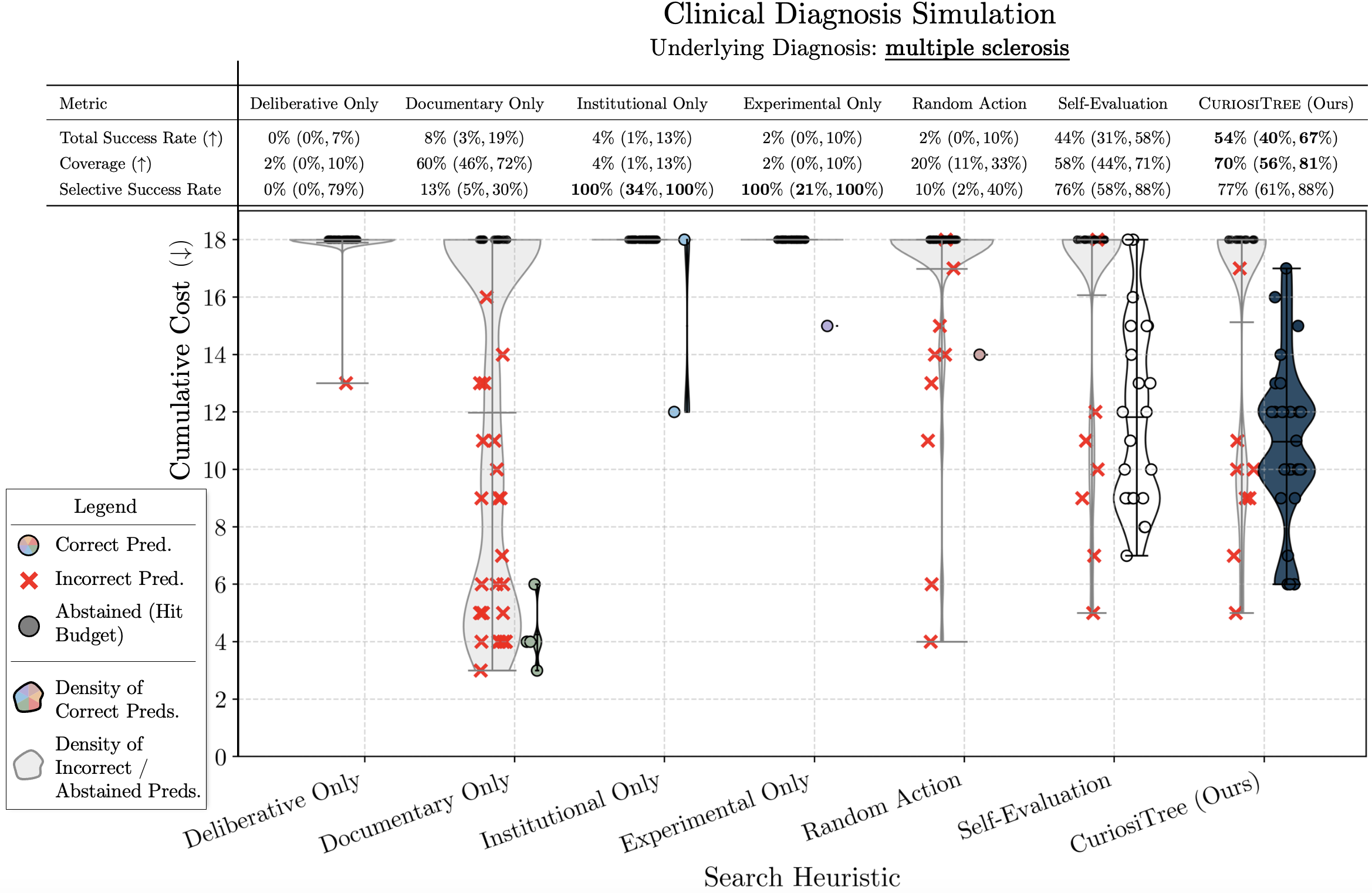}
    \caption{Result of simulation runs with a ground-truth diagnosis ``multiple sclerosis''. This figure is interpreted the same way as Figure \ref{fig:main-result}.}
\end{figure}

\begin{figure}[H]
    \centering
    \includegraphics[width=0.95\textwidth]{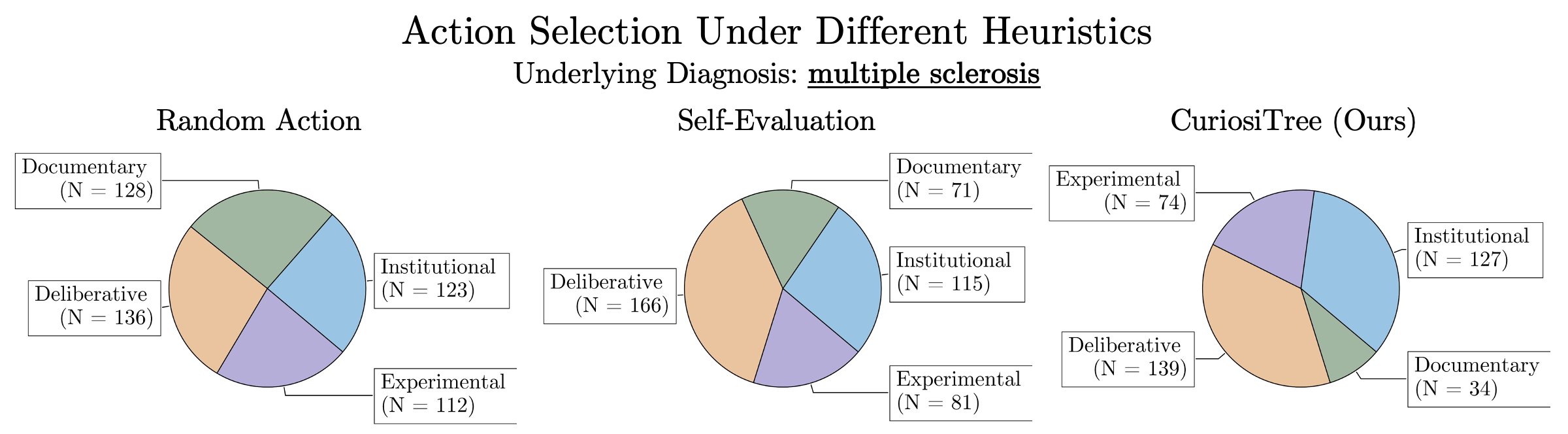}
    \caption{The action classes selected under different policies when diagnosing ``multiple sclerosis''. This figure is interpreted the same way as Figure \ref{fig:action-selection}.}
\end{figure}

\begin{dialoguebox}{Successfully Diagnosing ``Multiple Sclerosis'' with \textsc{CuriosiTree}}{A sequence of actions and responses leading to the successful diagnosis of ``multiple sclerosis'' using \textsc{CuriosiTree}.}

\question{What seems to be the main reason you've come to see me today?}\\
\environment{The patient responds, I've been having some trouble with my vision and balance, and I've been feeling really tired lately.}

\medskip

\experiment{Magnetic Resonance Imaging of the Brain to evaluate for structural abnormalities or lesions affecting vision and balance}\\
\environment{The test yields: 
\begin{itemize}
  \item T1-weighted images show multiple periventricular and subcortical white matter lesions, with 5 lesions demonstrating enhancement after gadolinium administration.
  \item T2-weighted images reveal a total of 12 hyperintense lesions in the periventricular, subcortical, and infratentorial regions.
  \item FLAIR sequence shows multiple hyperintense lesions in the supratentorial and infratentorial regions, with involvement of the optic radiations and the middle cerebellar peduncles.
  \item No evidence of acute hemorrhage, mass effect, or significant atrophy.
\end{itemize}}

\medskip

\reasoning{We know that the patient's symptoms and MRI findings are not suggestive of an acute or infectious process. The presence of enhancing and non-enhancing lesions suggests a chronic and relapsing condition such as multiple sclerosis.}\\
\environment{N/A}

\medskip

\reasoning{We know that the presence of both enhancing and non-enhancing lesions on the patient's MRI suggests a chronic and relapsing condition}\\
\environment{N/A}

\medskip

\prediction{Is it Multiple Sclerosis?}\\
\textbf{[END -- success]}

\end{dialoguebox}

\subsubsection{Results for ``Cystic Fibrosis''}

\begin{figure}[H]
    \centering
    \includegraphics[width=\textwidth]{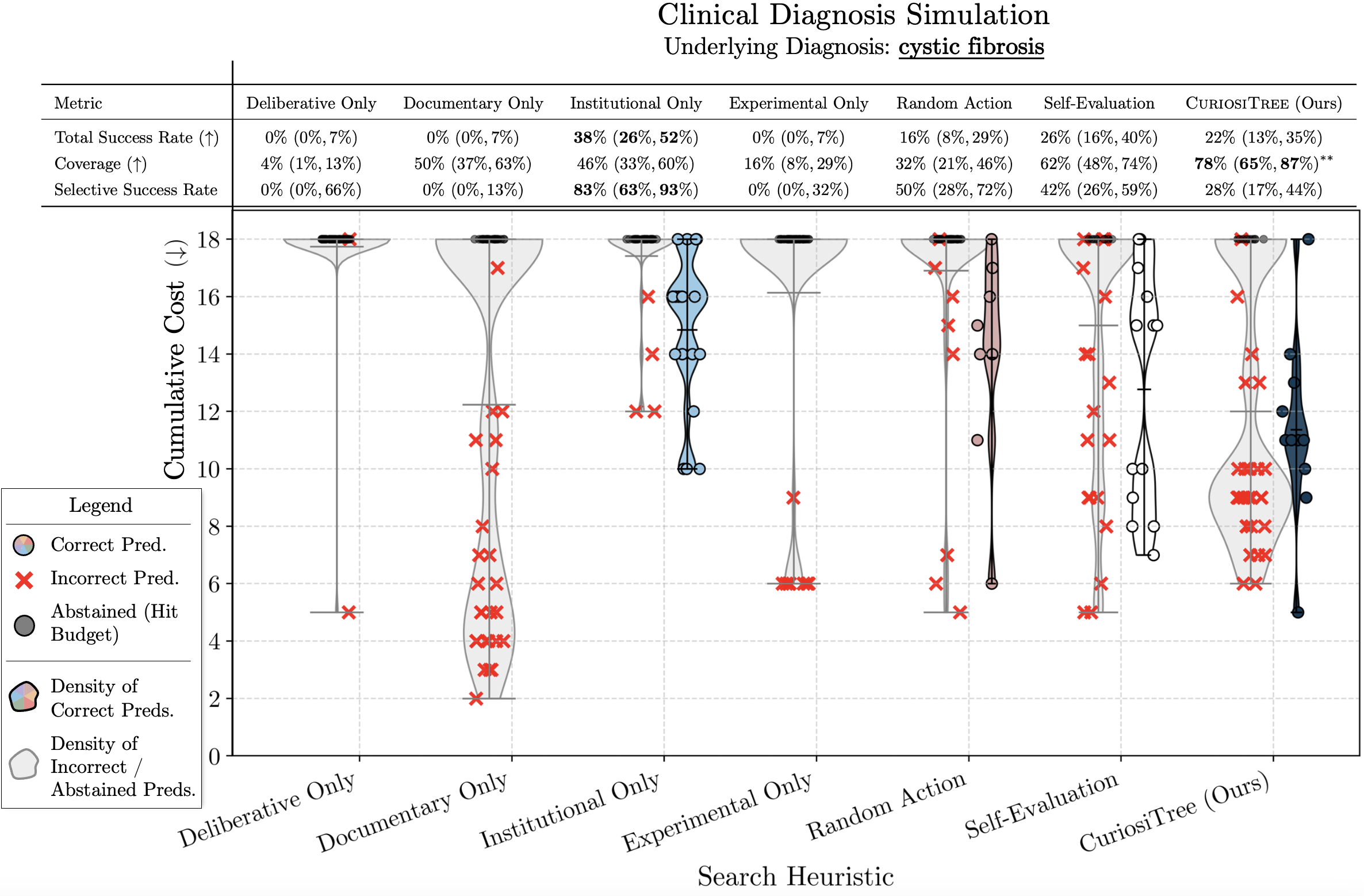}
    \caption{Result of simulation runs with a ground-truth diagnosis ``cystic fibrosis''. This figure is interpreted the same way as Figure \ref{fig:main-result}.}
\end{figure}

\begin{figure}[H]
    \centering
    \includegraphics[width=0.95\textwidth]{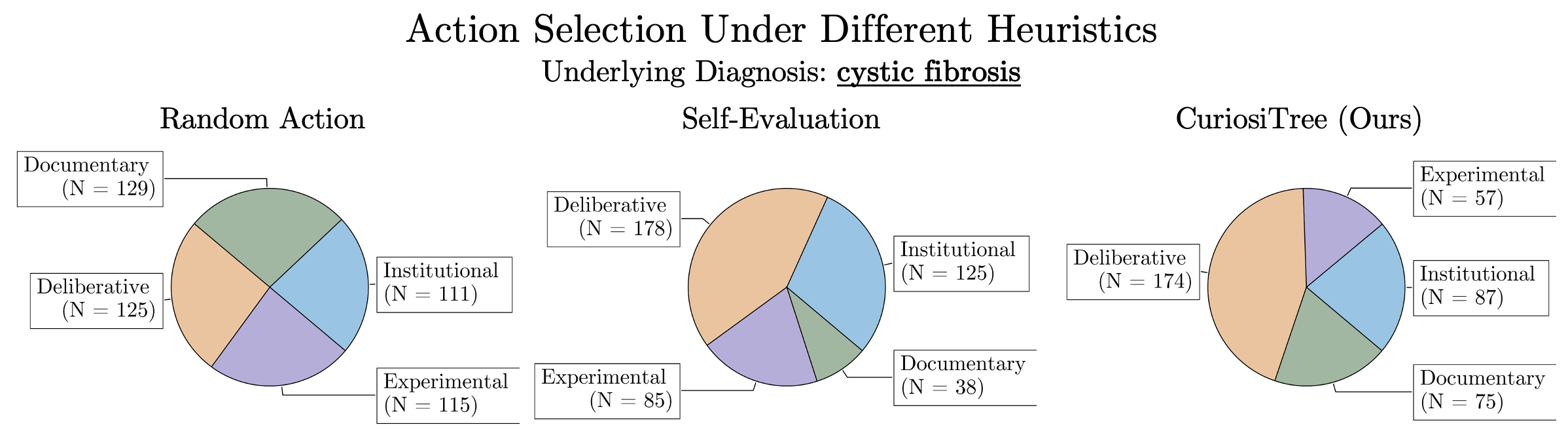}
    \caption{The action classes selected under different policies when diagnosing ``cystic fibrosis''. This figure is interpreted the same way as Figure \ref{fig:action-selection}.}
\end{figure}

\begin{dialoguebox}{Unsuccessfully Diagnosing ``Cystic Fibrosis'' with \textsc{CuriosiTree}}{A sequence of actions and responses attempting to diagnose ``cystic fibrosis'' using \textsc{CuriosiTree}.}

\question{What brings you to the clinic today?}\\
\environment{The patient responds, I've been having some trouble breathing and I've had a lot of lung infections lately.}

\medskip

\rag{Search Query: Causes of recurrent lung infections}\\
\environment{Retrieval: [Post-tuberculosis lung disease] Post-tuberculosis lung disease (PTLD) is ongoing lung disease that is caused by tuberculosis (TB) but persists after the infection is cured. PTLD can affect the airways, lung parenchyma, pulmonary vasculature, and pleura.}

\medskip

\experiment{Pulmonary Function Test to assess lung function and detect any obstructive or restrictive patterns}\\
\environment{The test yields: 
\begin{itemize}
\item Forced Expiratory Volume in 1 second (FEV1): 1.8 L (55\% predicted)
\item Forced Vital Capacity (FVC): 2.5 L (65\% predicted)
\item FEV1/FVC ratio: 0.72
\item Residual Volume (RV): 1.2 L (120\% predicted)
\item Total Lung Capacity (TLC): 4.2 L (85\% predicted)
\end{itemize}}

\medskip

\rag{Search Query: Chronic lung disease causes and symptoms}\\
\environment{Retrieval: [Chronic obstructive pulmonary disease] Chronic obstructive pulmonary disease (COPD) is a type of progressive lung disease characterized by chronic respiratory symptoms and airflow limitation. GOLD 2024 defines COPD as a heterogeneous lung condition characterized by chronic respiratory symptoms (dyspnea, cough, sputum production, or exacerbations) due to abnormalities of the airways (bronchitis, bronchiolitis) or alveoli (emphysema) that cause persistent, often progressive, airflow obstruction.}

\medskip

\reasoning{We know that the patient's Pulmonary Function Test results show a decreased FEV1/FVC ratio and increased Residual Volume. These findings are consistent with a diagnosis of chronic obstructive pulmonary disease.}\\
\environment{N/A}

\medskip

\prediction{Is it Chronic obstructive pulmonary disease?}\\
\textbf{[END -- failure]}

\end{dialoguebox}

\subsubsection{Results for ``Amyotrophic Lateral Sclerosis''}

\begin{figure}[H]
    \centering
    \includegraphics[width=\textwidth]{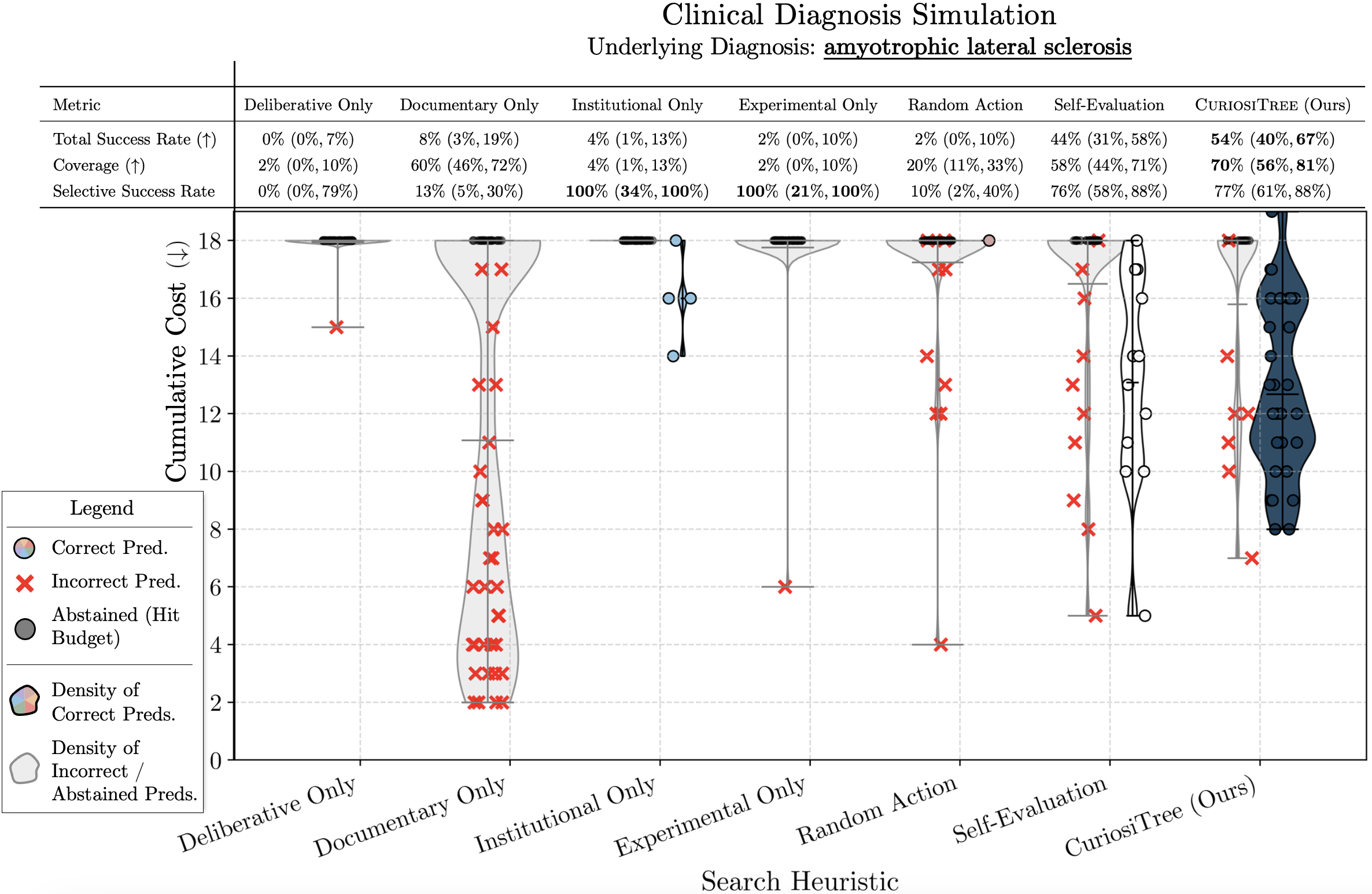}
    \caption{Result of simulation runs with a ground-truth diagnosis ``amyotrophic lateral sclerosis''. This figure is interpreted the same way as Figure \ref{fig:main-result}.}
\end{figure}

\begin{figure}[H]
    \centering
    \includegraphics[width=0.95\textwidth]{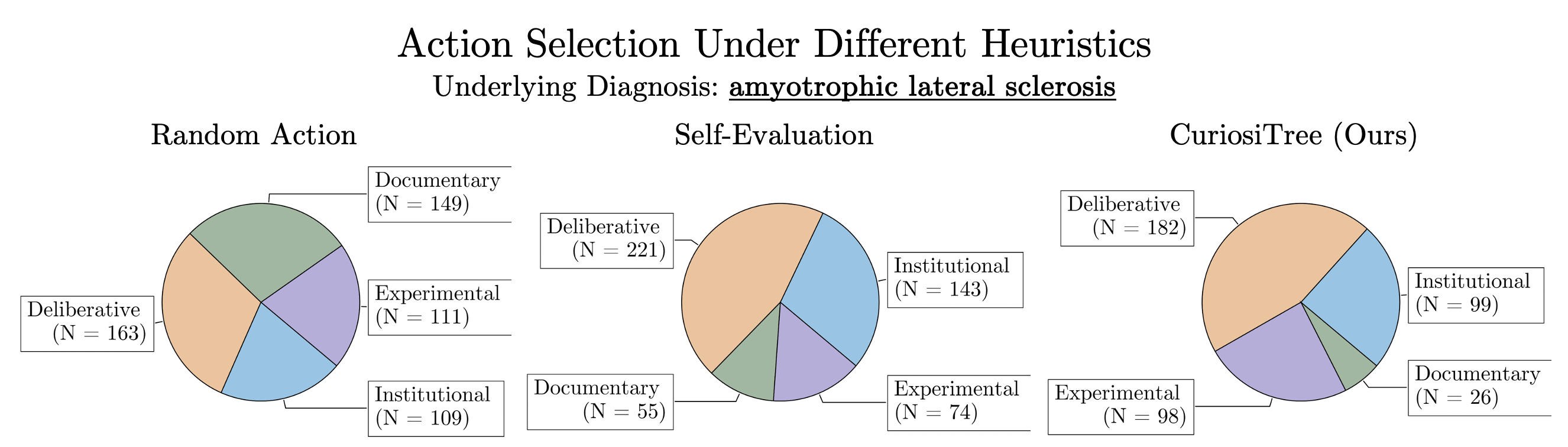}
    \caption{The action classes selected under different policies when diagnosing ``amyotrophic lateral sclerosis''. This figure is interpreted the same way as Figure \ref{fig:action-selection}.}
\end{figure}

\begin{dialoguebox}{Successfully Diagnosing ``Amyotrophic Lateral Sclerosis'' with \textsc{CuriosiTree}}{A sequence of actions and responses leading to the successful diagnosis of ``amyotrophic lateral sclerosis'' using \textsc{CuriosiTree}.}

\question{What seems to be the main reason you're visiting the clinic today?}\\
\environment{The patient responds, I've been having some trouble with my arm, it's been feeling weak and I'm having a hard time holding things.}

\medskip

\experiment{Creatine Kinase level to evaluate muscle damage or disease}\\
\environment{The test yields: 543 U/L.}

\medskip

\experiment{Muscle Biopsy to examine muscle tissue for signs of disease or damage}\\
\environment{The test yields: Muscle biopsy shows denervation atrophy, with groups of atrophic fibers and scattered angulated fibers, and evidence of reinnervation, including fiber type grouping.}

\medskip

\reasoning{which is consistent with a motor neuron disease or peripheral neuropathy}\\
\environment{N/A}

\medskip

\reasoning{We know that the patient's elevated creatine kinase level and muscle biopsy results suggest an ongoing process of muscle damage and repair}\\
\environment{N/A}

\medskip

\reasoning{We know that the patient's elevated creatine kinase level and muscle biopsy results suggest an ongoing process of muscle damage and repair}\\
\environment{N/A}

\medskip

\reasoning{We know that the combination of muscle weakness and denervation atrophy suggests a disease affecting the motor neurons}\\
\environment{N/A}

\medskip

\question{Do you have difficulty swallowing or speaking?}\\
\environment{The patient responds, "Sometimes I have trouble getting words out, and swallowing food can be a bit of a problem, especially with thicker foods."}

\medskip

\reasoning{We know that the patient's symptoms and laboratory results are suggestive of a specific type of motor neuron disease that affects both upper and lower motor neurons}\\
\environment{N/A}

\medskip

\prediction{Is it Amyotrophic Lateral Sclerosis?}\\
\textbf{[END -- success]}

\end{dialoguebox}

\subsubsection{Results for ``Myasthenia Gravis''}

\begin{figure}[H]
    \centering
    \includegraphics[width=\textwidth]{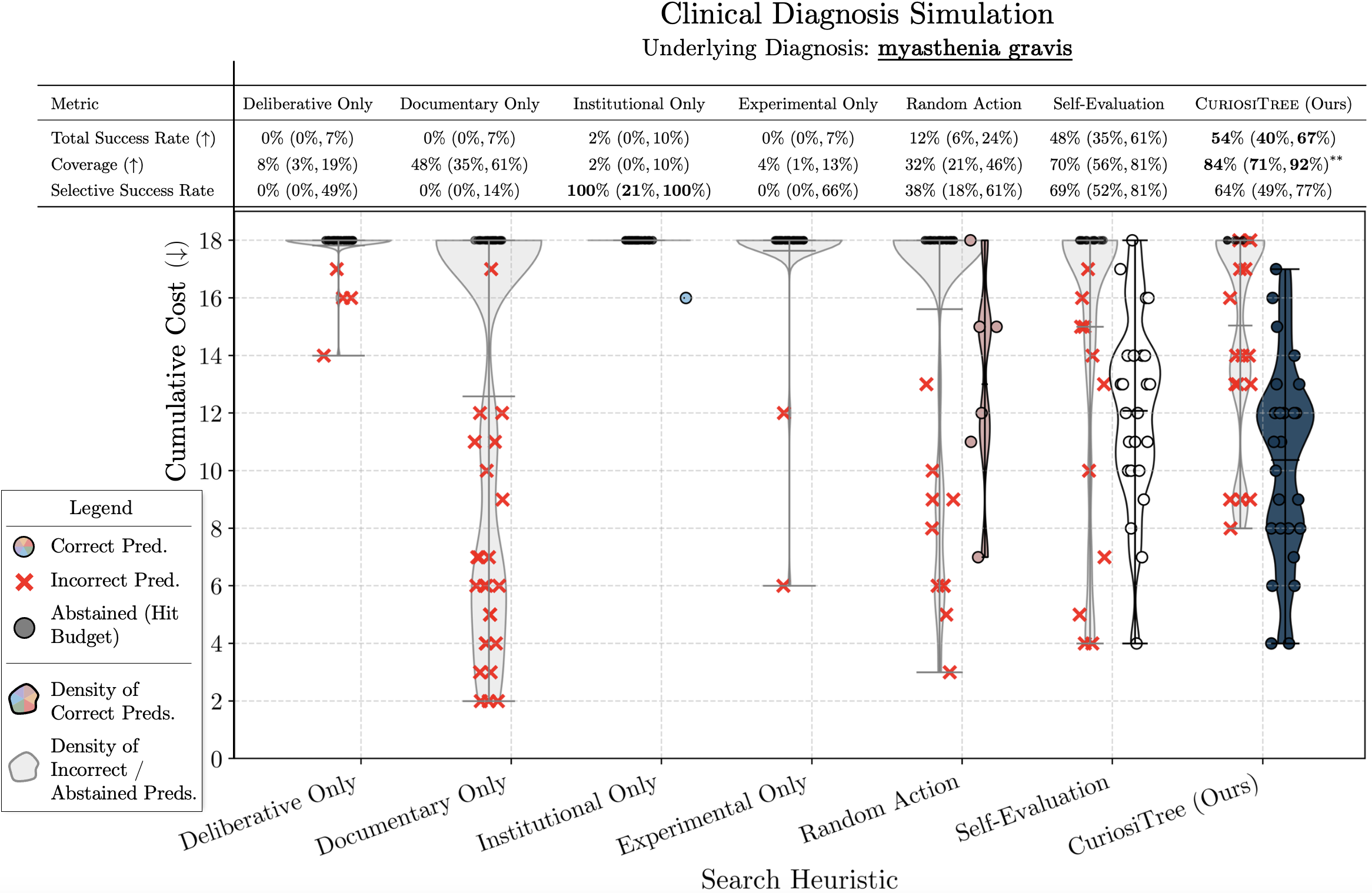}
    \caption{Result of simulation runs with a ground-truth diagnosis ``myasthenia gravis''. This figure is interpreted the same way as Figure \ref{fig:main-result}.}
\end{figure}

\begin{figure}[H]
    \centering
    \includegraphics[width=0.95\textwidth]{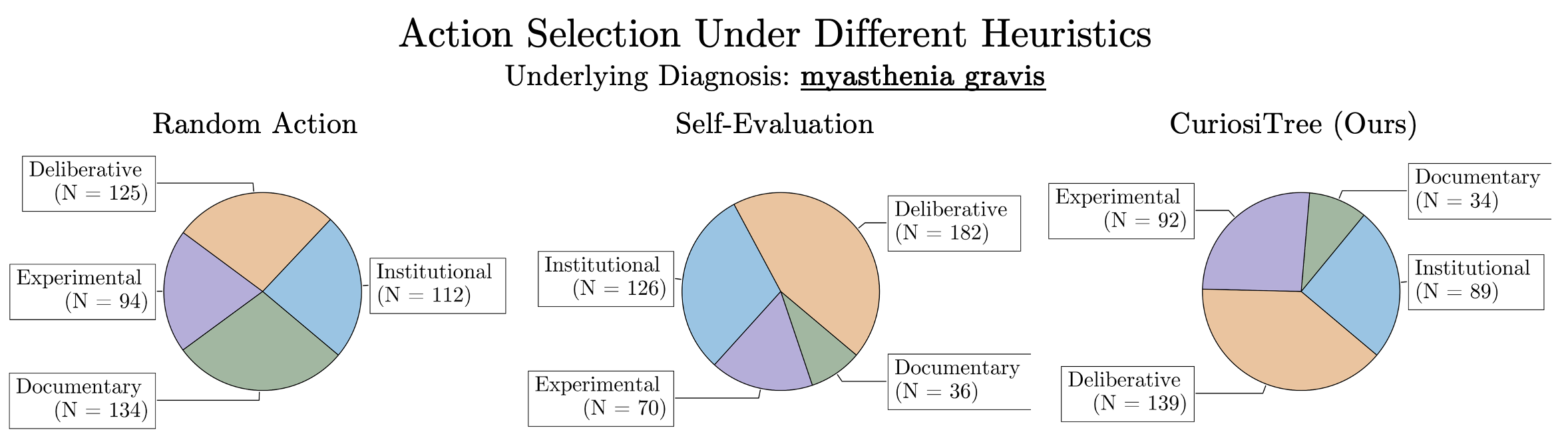}
    \caption{The action classes selected under different policies when diagnosing ``myasthenia gravis''. This figure is interpreted the same way as Figure \ref{fig:action-selection}.}
\end{figure}

\begin{dialoguebox}{Successfully Diagnosing ``Myasthenia Gravis'' with \textsc{CuriosiTree}}{A sequence of actions and responses leading to the successful diagnosis of ``myasthenia gravis'' using \textsc{CuriosiTree}.}

\question{What brings you to the clinic today?}\\
\environment{The patient responds, I've been having some trouble with my muscles feeling weak, especially after I've been active or at the end of the day.}

\medskip

\experiment{Initialized by: Dr. [Last Name] Date: [Current Date] Requested Test: Electromyography to assess the electrical activity of the muscles and diagnose conditions such as myasthenia gravis or amyotrophic lateral sclerosis}\\
\environment{The test yields: 
\begin{itemize}
\item Decreased compound muscle action potential (CMAP) amplitudes with repetitive nerve stimulation (RNS)
\item A 30\% decrement in CMAP amplitude at 3 Hz stimulation of the ulnar nerve
\item Single-fiber electromyography (SFEMG) showing increased jitter (mean consecutive difference, MCD: 55 $\mu$s) and blocking (11\% of fibers)
\end{itemize}}

\medskip

\prediction{Is it Myasthenia Gravis?}\\
\textbf{[END -- success]}

\end{dialoguebox}
\end{document}